\documentclass[runningheads]{llncs}
\usepackage{graphicx}
\usepackage{tikz}
\usepackage{comment}
\usepackage{amsmath,amssymb} % define this before the line numbering.
\usepackage{color}
\usepackage[accsupp]{axessibility}  % Improves PDF readability for those with disabilities.

\usepackage{siunitx}

\usepackage[para,online,flushleft]{threeparttable}
\usepackage{bm}
\usepackage{subcaption}

\usepackage[pagebackref,breaklinks,colorlinks]{hyperref}
\usepackage[normalem]{ulem}

\usepackage[normalem]{ulem}
\usepackage{xcolor}
\usepackage[figuresright]{rotating}
\usepackage{enumitem}
\setitemize{noitemsep,topsep=0pt,parsep=0pt,partopsep=0pt}

\usepackage{booktabs}
\usepackage{multirow,multicol,makecell}
\usepackage{pifont}
\usepackage{caption}
\newcommand*\samethanks[1][\value{footnote}]{\footnotemark[#1]}

\begin{document}
% \linenumbers
\pagestyle{headings}
\mainmatter
\def\ECCVSubNumber{100}  % Insert your submission number here

\title{SPViT: Enabling Faster Vision Transformers via Latency-aware Soft Token Pruning} % Replace with your title

\begin{comment}
\titlerunning{ECCV-22 submission ID \ECCVSubNumber} 
\authorrunning{ECCV-22 submission ID \ECCVSubNumber} 
\author{Anonymous ECCV submission}
\institute{Paper ID \ECCVSubNumber}
\end{comment}
%******************

%\begin{comment}
\titlerunning{SPViT}
\author{Zhenglun Kong\thanks{Both authors contributed equally.} \inst{1} \and
Peiyan Dong\samethanks \inst{1} \and
Xiaolong Ma\inst{2}  \and
Xin Meng\inst{3}  \and
Wei Niu\inst{4} \and
Mengshu Sun\inst{1} \and
Xuan Shen\inst{1} \and
Geng Yuan\inst{1}  \and
Bin Ren\inst{4} \and
Hao Tang\inst{5}  \and
Minghai Qin\inst{1}  \and
Yanzhi Wang\inst{1} 
}
\authorrunning{Z. Kong, P. Dong et al.}
\institute{Northeastern University, Boston MA 02115, USA \\
\email{\{kong.zhe,dong.pe,yanz.wang\}@northeastern.edu} \and
Clemson University, Clemson SC 29634, USA \and
Peking university, Beijing 100871, China \and
College of William and Mary, Williamsburg  VA 23185, USA \and
%Western Digital Research \and
CVL, ETH Zürich, Zürich 8092, Switzerland
}
%\end{comment}
%******************

\maketitle

\begin{abstract}
Recently, Vision Transformer (ViT) has continuously established new milestones in the computer vision field, while the high computation and memory cost makes its propagation in industrial production difficult. Considering the computation complexity, the internal data pattern of ViTs, and the edge device deployment, we propose a latency-aware soft token pruning framework, \textbf{SPViT}, which can be set up on vanilla Transformers of both flatten and hierarchical structures, such as DeiTs and Swin-Transformers (Swin).
More concretely, we design a dynamic attention-based multi-head token selector, which is a lightweight module for adaptive instance-wise token selection.
% We further introduce a soft pruning technique, which integrates the less informative tokens generated by the selector module into a package token that will participate in subsequent calculations rather than being completely discarded. 
We further introduce a soft pruning technique, which integrates the less informative tokens chosen by the selector module into a package token rather than discarding them completely. 
SPViT is bound to the trade-off between accuracy and latency requirements of specific edge devices through our proposed latency-aware training strategy.
Experiment results show that SPViT significantly reduces the computation cost of ViTs with comparable performance on image classification. 
Moreover, SPViT can guarantee the identified model meets the latency specifications of mobile devices and FPGA, and even achieve the real-time execution of DeiT-T on mobile devices. 
For example, SPViT reduces the latency of DeiT-T to 26 ms (26\%$\sim $41\% superior to existing works) on the mobile device with 0.25\%$\sim $4\%  higher top-1 accuracy on ImageNet. Our code is released
at \url{https://github.com/PeiyanFlying/SPViT}
%We provide our code in the supplementary.
\end{abstract}

\section{Introduction}
\begin{figure}[htb]
\centering
\includegraphics[width=0.7\columnwidth]{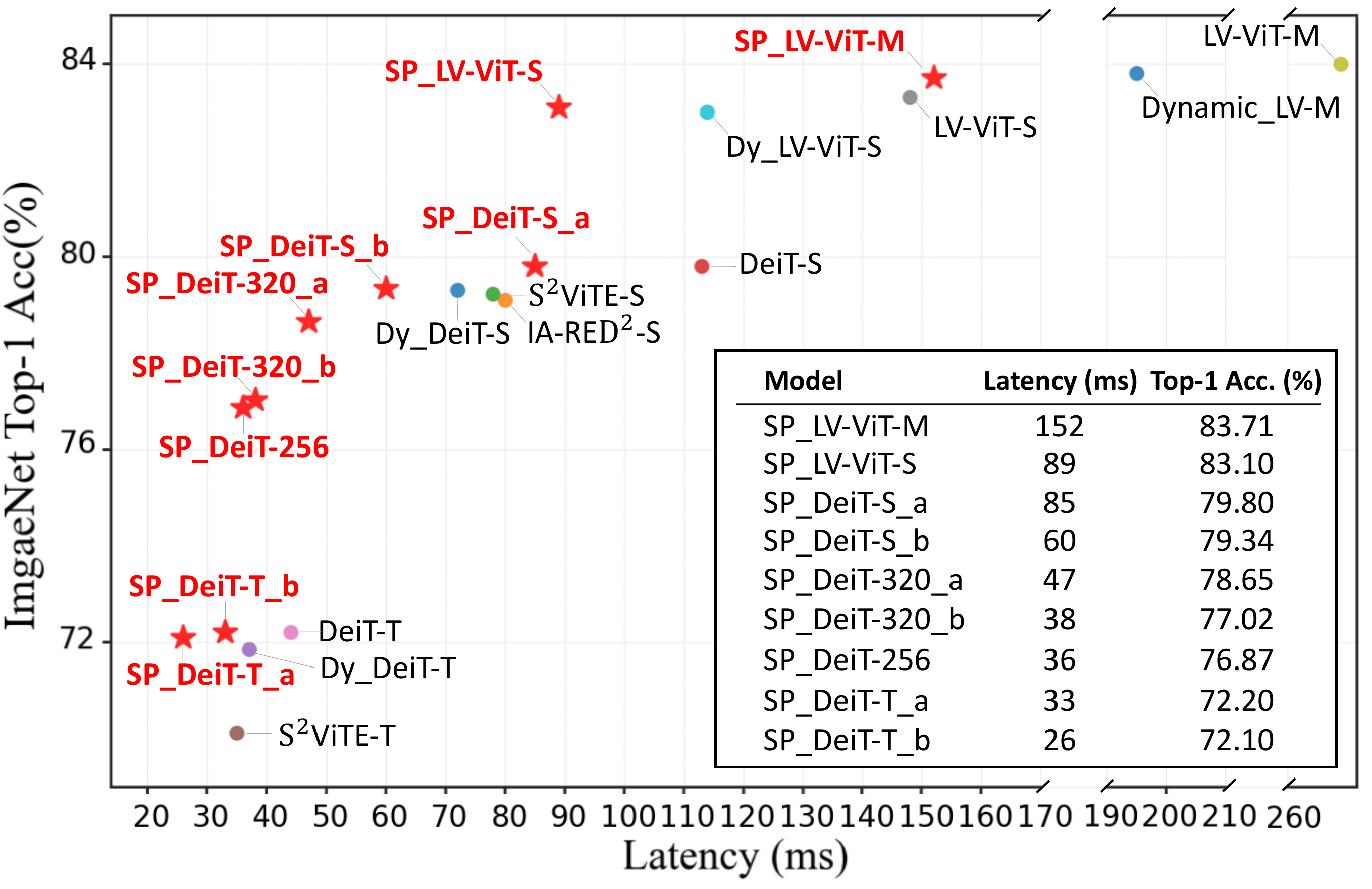}
\caption{Comparison of different pruning methods with various accuracy-latency trade-offs. %We can reduce the latency of larger models with negligible accuracy decrease. Furthermore, our models outperform lightweight models under similar latency. 
We can increase the accuracy of light weight models at similar latency, and expedite larger models with negligible decrease of accuracy.
Models are tested on Samsung Galaxy S20.}
\label{fig:main_latency}
%%%\vspace{-0.4cm}
\end{figure}

\sloppy
Recently, a new trend of leveraging Transformer architecture~\cite{vaswani2017attention} into the computer vision domain has emerged~\cite{hudson2021ganformer,chen2021pix2seq,kim2021hotr,Deng_2021_ICCV,xue2021transfer,zhao2021point,Guo_2021,srinivas2021bottleneck}. 
The Vision Transformer (ViT), which solely exploits the self-attention mechanism that inherits from the Transformer architecture, has set up many state-of-the-art (SOTA) records in image classifications~\cite{dosovitskiy2021an,Touvron2021TrainingDI,chen2021crossvit}, object detection~\cite{carion2020end,dai2021up,amini2021t6d,misra2021-3detr}, tracking~\cite{chen2021transformer,yan2021learning,meinhardt2021trackformer}, semantic segmentation~\cite{zheng2021rethinking,cheng2021perpixel}, depth estimation~\cite{yang2021transformers,li2021revisiting}, image retrieval~\cite{el2021training}, and image enhancement~\cite{yang2020learning,chen2021pre,lu2021efficient}.
However, despite the impressive general results, ViTs have sacrificed lightweight model capacity, portability, and trainability in return for high accuracy. The mass amount of computations brought by operations (e.g. Conv, MatMul, Add) in existing models remains a setback for edge device deployment.

Pruning has been proved as the one of the most effective methods to reduce network dimensions in convolution-based neural networks~\cite{ren2019admm,yuan2021mest,ma2021sanity,liu2021lottery,zhang2021unified,niu2021grim,rumi2020accelerating,ma2021non,zhang2021structadmm,chen2022coarsening,hou2022chex,ma2022blcr,ma2021effective,chang2021mix}. However, when huge amount of AI-powered applications are benefiting from the network pruning advantages~\cite{niu2020patdnn,ma2020pconv,ma2020image,yuan2021improving,yuan2021tinyadc,gong2020privacy,tan2020pcnn,chu2020pim,ma2020tiny,yuan2019ultra,li2022neural,10.5555/3491440.3491828,fang2020encoding,fang2021neuromorphic}, the applications of self-attention-based neural network pruning remain scarce~\cite{guo2020accelerating,sanh2020movement,li2020efficient,wang2021spatten,niu2021compression}. 
There still exists a gap between the actual device deployment and acceleration in the ViT pruning frameworks. For instance, attention head pruning~\cite{chen2021chasing} performs weight pruning on the transformation matrix ($W_Q$, $W_K$, $W_V$) before the multi-head self-attention (MSA) operation. It is an inefficient way for computation reduction because only part of the ViT computations (i.e., MSA) can be alleviated (see Sec. \ref{complex_anal} for justification).
In a lightweight model, head pruning cannot guarantee an ideal pruning rate without significant accuracy deterioration.
Static token pruning~\cite{rao2021dynamicvit} reduces the number of input tokens by a fixed ratio for different images, which restricts the image pruning rate, ignoring the fact that the high-level information of each image varies both in the region size and location.
Furthermore, it is difficult for the deployment on edge devices since newly introduced operations (e.g., Argsort) are currently not well supported by many frameworks~\cite{prillo2020softsort}.
In contrast, dynamic token pruning~\cite{pan2021iared2} deletes redundant tokens based on the inherent image characteristics to achieve a per-image adaptive pruning rate.
However, this method implies a potentially huge search space, which will easily cause a limited overall pruning rate or undermined accuracy if the token selection mechanism is not carefully designed. %\cross{different images with different pruning rates}
In addition, the pruning mechanism in~\cite{pan2021iared2} unreservedly discards less informative tokens, which results in the loss of the informative part of the removed tokens.

\begin{figure*}[t!]
\centering
\includegraphics[width=0.9\columnwidth]{ 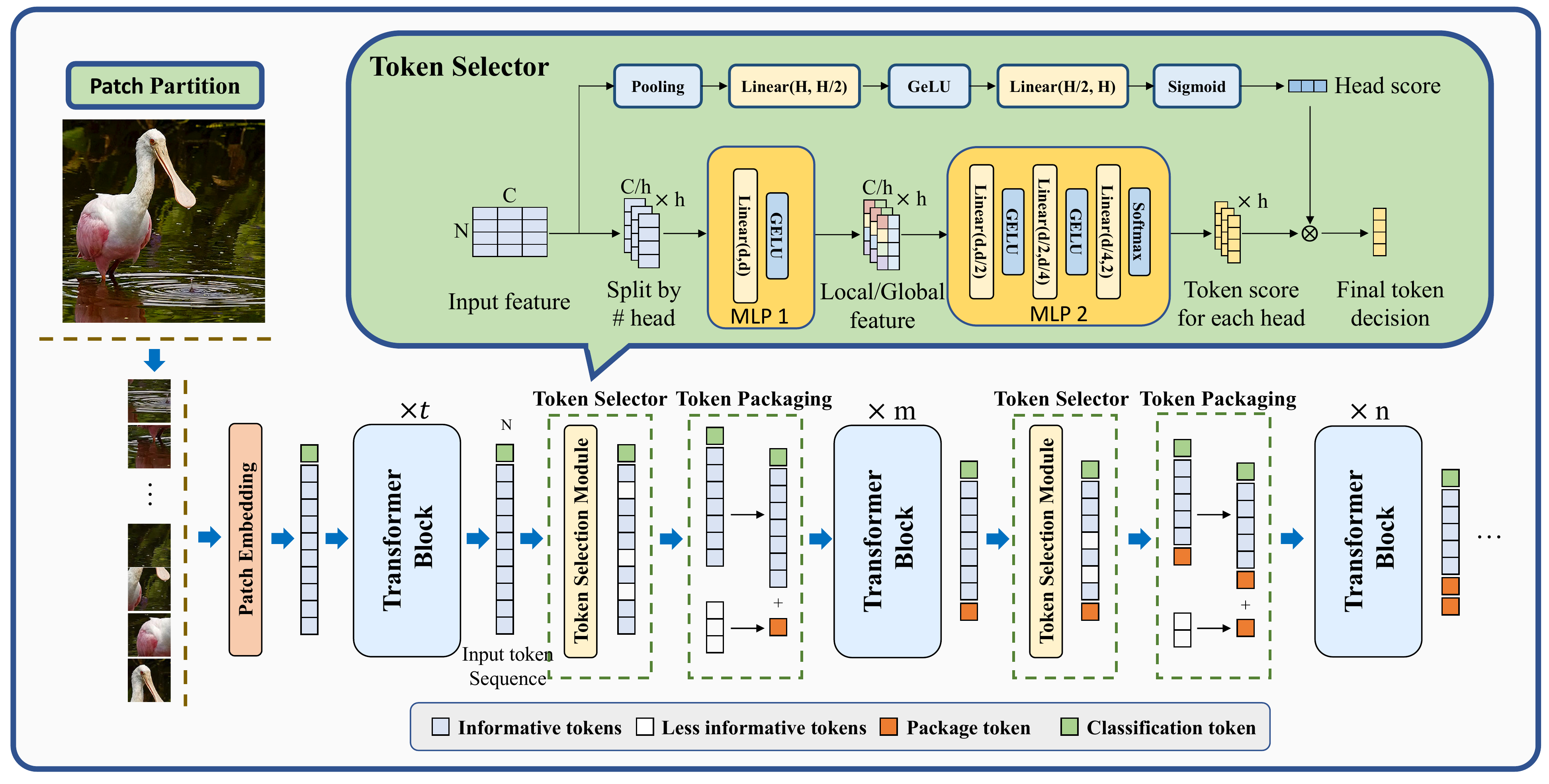}
\caption{Overall workflow. Bottom figure: Token selector is inserted multiple times throughout the model, along with the token packaging technique to generate a package token from the less informative tokens. The package token is concatenated with the informative tokens to be fed in the following transformer blocks. Upper figure: Our attention-based multi-head token selector to obtain token scores for keep/prune decisions. }
\label{fig:framework}
%%%\vspace{-0.4cm}
\end{figure*}

In this paper, we manage to overcome the above limitations.
Specifically, as shown in Fig.~\ref{fig:framework}, we propose a latency-aware \underline{S}oft \underline{P}runing framework (SPViT), which simultaneously optimizes ViT accuracy and maximizes per-image dynamic pruning rate while maintaining actual computation constraints on edge devices.
In ViT, each head encodes the visual receptive field independently~\cite{pan2021iared2,heo2021pit,mao2021dual}, which implies that each token has a different influence in different heads~\cite{dosovitskiy2021an,zhai2021scaling,yu2021glance,gao2021container}.
We thus propose a token selector to evaluate the importance score of each token based on its characteristic statistics in all heads. Then, through an attention-based branch~\cite{hu2018squeeze} in the selector, we calculate the weighted sum of each score to obtain the final score of a token, which determines whether the token should be pruned.
With the token selector, all tokens generated from the input images can be precisely ranked and pruned based on their importance scores and thus achieving a high overall pruning rate.

The token representations~\cite{wu2020visual,xu2021you,chen2021exploring,chefer2021transformer} in early and middle layers are insufficiently encoded, which makes token pruning quite difficult.
To mitigate the challenge, we introduce a package token technique, which compresses the less-informative tokens, picked out by the token selector, into a package token. Then, we concatenate the package token to the remaining tokens for subsequent blocks.
On the one hand, although informative tokens may be discarded due to the poor encoding ability in earlier blocks of ViT~\cite{xu2022evovit}, this error will be partly corrected by the residual information stored in the package token.
On the other hand, background features can help emphasize foreground features~\cite{yang2021instance}. Completely removing less informative (negative) tokens will weaken the ability of self-attention to capture key information. Therefore, the package token can serve as a way to help preserve background features.
By adding minimal computation cost, the token pruning rate can be increased significantly.

% Taking the hardware efficiency into consideration, all the operators contained in our framework have been well-supported on edge devices.
In addition, we elaborate a latency-aware training strategy, which consists of two parts: latency-aware loss function and layer-to-phase progressive training.
The former bridges the token pruning rates with latency specifications of diverse edge devices.
The latter indicates that we progressively insert one selector in each block and train the new selector under the latency budget of the target device.
Next, we group adjacent blocks with similar pruning rates into a phase, keep the first selector in this phase and remove others.
While maintaining high accuracy, it can search for the appropriate pruning rate for each block and the desired insertion position of the selector. Fig.~\ref{fig:main_latency} shows the on device performance of our model compared with other pruned or scaled models.

Our contributions are summarized as follows:
\begin{itemize}[leftmargin=*, noitemsep,topsep=0pt]
 \item We provide a detailed analysis on the computational complexity of ViT and different compression strategies.
 Based on our analysis, token pruning holds a greater computation reduction compared to the compression of other dimensions.
\item Considering the vision pattern inside ViT, we propose SPViT, a novel  method which includes the attention-based multi-head token selector and the token packaging technique to achieve per-image adaptive pruning. We design a latency-aware training strategy, which efficiently explores the SPViT design space given the hardware latency budget, and maximizes the per-image pruning rate without any accuracy degradation.
 \item SPViT enables a higher pruning rate than other state-of-the-art with comparable accuracy.
 For lightweight models, SPViT allows the DeiT-S and DeiT-T to reduce inference latency by 40\%-60\% within 0.5\% accuracy loss. 
 It can further generate more efficient PiTs and Swins with negligible performance drops.
 In particular, SPViT is superior in the compression of lightweight models.
 \item We demonstrate a real-time realization of DeiT-T on mobile phones (e.g., 26 $ms$ on a Samsung Galaxy S20) and DeiT-S on a Xilinx FPGA (13.2 $ms$ on a Xilinx ZCU102).
 To the best of our knowledge, it is the first time that the ViT models perform inference on the edge devices beyond real-time\footnote{Real-time inference usually means 30 frames per second, which is approximately 33 $ms$/image.}.
\end{itemize}

\section{Related Work}
\label{gen_inst}
\noindent\textbf{Vision Transformers}. ViT~\cite{dosovitskiy2021an} is a pioneering work that uses only a Transformer to solve various vision tasks. Compared to traditional CNN structures, ViT allows all the positions in an image to interact through transformer blocks, whereas CNNs operate on a fixed-sized window with restricted spatial interactions, which can have trouble capturing relations at the pixel level in both spatial and time domains~\cite{raghu2021vision}. 
Since then, many variants have been proposed~\cite{Graham_2021_ICCV,liu2021Swin,Yuank_2021_ICCV,wang2021pyramid,han2021transformer,wu2021rethinking,chen2021autoformer,steiner2021train,el-nouby2021xcit,liu2021efficient,wang2021kvt,bao2022beit}. For example, DeiT~\cite{Touvron2021TrainingDI}, T2T-ViT~\cite{Yuan_2021_ICCV} and Mixer~\cite{chen2022when} tackle the data-inefficiency problem in ViT by training only with ImageNet. PiT~\cite{heo2021pit} replaces the uniform structure of Transformer with depth-wise convolution pooling layer to reduce spacial dimension and increase channel dimension. LV-ViT~\cite{jiang2021all} introduces a token labeling 
method to improve training. PS-ViT~\cite{Yue_2021_ICCV} %abandons the fixed length tokens with 
applied progressive sampled tokens.

\noindent\textbf{Efficient ViT.}
The huge memory usage and computation cost of the self-attention mechanism serve as the roadblock to the efficient deployment of ViT models on edge devices. Many works aim at accelerating the inference speed of ViT~\cite{chen2021psvit}.
For instance, S${^2}$ViTE~\cite{chen2021chasing} prunes token and attention head in a structured way via sparse training.
VTP~\cite{zhu2021visual} reduces the input feature dimension by learning their associated importance scores with L1 regularization. 
IA-RED${^2}$ ~\cite{pan2021iared2} drops redundant tokens with a multi-head interpreter.
PS-ViT (T2T)~\cite{tang2021patch} discards useless patches in a top-down paradigm.
DynamicViT~\cite{rao2021dynamicvit} removes redundant tokens by estimating their importance score with a MLP~\cite{vaswani2017attention} based prediction module.
Evo-ViT~\cite{xu2022evovit} develops a slow-fast token evolution method to preserve more image information during pruning.
TokenLearner~\cite{ryoo2021tokenlearner} and PATCHMERGER~\cite{renggli2022learning} uses spatial attention to generate a small set of token vectors adaptive to the input.
However, to the best of our knowledge, our idea of considering actual edge device deployment and acceleration has not been investigated by any existing ViT pruning methods.

\section{Computational Complexity Analysis}
\label{complex_anal}

\begin{table*}[t] \small
\caption{The computational complexity of each operation in a ViT block. The input $N{\times} D_{ch}$ goes through three linear transformation layers with $D_{ch}{\times} D_{attn}$ to generate Query ($Q$), Key ($K$), and Value ($V$) matrices of size $N{\times} D_{attn}$. $N$ is transitive, while $D_{ch}$ is not.}
\label{complex_chart}
\small
\centering
\scalebox{0.75}{
\begin{tabular}{c | c | c | c | c | c | c}
\toprule
\# & Module & Input Size & Operation & Layer Size & Output Size & Computation \\ \midrule
\ding{172} & \multirow{5}{*}{MSA} & $N \times D_{ch}$ & Linear Transformation & $D_{ch}\times D_{attn}$ & $N \times D_{attn}$ & $ND_{ch}D_{attn}\times 3$ \\ \cmidrule{1-1} \cmidrule{3-7}
\ding{173} & & $N \times D_{attn}$ & $Q$ Multiplying $K^{T}$ & - & $N \times N$ & $N^{2}D_{attn}$ \\ \cmidrule{1-1} \cmidrule{3-7}
\ding{174} & & $N \times N$ & Multiplying $V$ & - & $N \times D_{attn}$ & $N^{2}D_{attn}$ \\ \cmidrule{1-1} \cmidrule{3-7}
\ding{175} &  & $N \times D_{attn}$ & Projection & $D_{attn}\times D_{ch}$ & $N \times D_{ch}$ & $ND_{attn}D_{ch}$ \\ \midrule
\ding{176} & \multirow{2}{*}{FNN} & $N \times D_{ch}$ & FC Layer & $D_{ch}\times 4D_{fc}$ & $N \times 4D_{fc}$ & $4ND_{ch}D_{fc}$  \\ \cmidrule{1-1} \cmidrule{3-7}
\ding{177} & & $N \times 4D_{fc}$ & FC Layer & $4D_{fc}\times D_{ch}$ & $N \times D_{ch}$ & $4ND_{fc}D_{ch}$ \\ \midrule
\multicolumn{6}{c|}{\multirow{2}{*}{Total Computational Complexity}}  & $4ND_{ch}D_{attn} + $ \\
\multicolumn{6}{c|}{}& $2N^{2}D_{attn} + 8ND_{ch}D_{fc}$ \\
\bottomrule
\end{tabular}}
%%%\vspace{-0.4cm}
\end{table*}

Given an input sequence $N{\times} D$, where $N$ is the input sequence length or the token number and $D$ is the embedding dimension~\cite{Touvron2021TrainingDI} of each token, some works~\cite{pan2021iared2,zhu2021visual} address the computational complexity of ViT as $(12ND^2{+}2N^2D)$. However, $D$ represents different dimensions and should be written as $(4ND_{ch}D_{attn}{+}2N^2D_{attn}{+}8ND_{ch}D_{fc})$. Neglecting the difference may cause misleading conclusions, especially when analyzing the validity of pruning methods such as token pruning and dimension pruning. 

Table~\ref{complex_chart} shows an analysis of each operation in a Transformer block. There are three main branches of ViT pruning.
(i) Token channel pruning: The sequence tokens are pruned along $D_{ch}$ dimension. $D_{ch}$ is non-transmissible, which means reducing input dimension only affects the computation of the current matrix multiplication. 
To reduce computation for all layers, a mask layer is added to multiply with the input before going through the linear layer~\cite{zhu2021visual}. 
(ii) Token pruning: $N$ is transitive, so directly pruning tokens will contribute to the linearly or even quadratically ($N^2$ in \ding{173} and \ding{174}) reduction of all operations. %\xl{[XL: all some details, why linear or exponential?]}
(iii) Attention head pruning (or attention channel pruning): The pruning operations are performed on weight tensors of each attention head in the MSA module. However, only the $D_{attn}$ in the MSA module can be counted towards computation reduction, which usually contributes less than 40\% of the total computation in most ViT architectures. 
Therefore, with the same pruning rate, pruning tokens (reducing $N$) can reduce more overall computation than pruning channels (reducing $D_{ch}$ or $D_{attn}$).

\section{Latency-Aware Soft Pruning}
In this section, we first introduce our soft token pruning framework. Then, we show an elaborate design of each module. Finally, we give a detailed discussion of our latency-aware training strategy. 

\subsection{Framework Overview}
Our soft pruning framework includes a token selector and a token packaging technique. We propose a hierarchical pruning scheme, where these two modules are inserted between multiple blocks throughout the model. 
As shown in Fig.~\ref{fig:framework}, the input token sequence first goes through a token selector, where each token is scored and defined as either informative or less informative. After that, less informative tokens are separated from the sequence and integrated into a package token. This package token then concatenates to the informative tokens to involve in subsequent calculations in the blocks. In the next phase, a newly generated package token will connect with the existing package token.

For ViT training with our framework, we devise a latency-aware sparsity loss for the hardware's maximum computation bandwidth. We perform a layer-to-phase progressive training schedule to compress the search space, where model accuracy optimization and hardware computation reduction can be simultaneously achieved. The overall framework is hardware friendly with no unsupported operations and miniature computation cost.  

% \subsection{Attention-based Multi-head Token Selector}
% \begin{figure}[t]
% \centering
% \includegraphics[width=0.9\columnwidth]{Figs/headattention.pdf}
% \caption{Heatmaps showing the informative region detected by each head in DeiT-S. Each attention head focuses on encoding different image features and visual receptive fields.}
% \label{fig:headattention}
% %\vspace{-0.4cm}
% \end{figure}

\noindent\textbf{Multi-head Token Selector.} 
We propose a fine-grained approach to evaluate token scores. As shown in Fig.~\ref{fig:headattention}, in ViT’s multi-head vision pattern, each head focus on encoding different features and respective fields of an image. This implies that the importance of each token towards each head is different.
Our multi-head selector generates a list of token scores for each head. Let one head dimension be $d {=} C/H$, where $C$ is the input dimension and $H$ is the number of head. We split the input $X {\in} \mathbb{R}^{N\times C}$ by the number of attention head into $\{ x_i\}^{H}_{i=1}{\in} \mathbb{R}^{N\times d}$, and obtain local $f_i^{local}$ and global $f_i^{global}$ features separately through an MLP layer with a pipeline of $LayerNorm {\rightarrow} Linear(d,d/2) {\rightarrow} GELU$:
%%%\vspace{-0.2cm}
\begin{equation}
f_i^{local} = {\rm MLP}(x_i)  \in \mathbb{R}^{N\times d/2},
%f_i^{loc} = {\rm GELU}({\rm Linear}({\rm LayerNorm}(x_i)))  \in \mathbb{R}^{N\times d/2},
\label{eq:mlp1}
\end{equation}
%%%\vspace{-0.4cm}
\begin{equation}
f_i^{global} = {\rm AvgPool}( {\rm MLP}(x_i),D)  \in \mathbb{R}^{1\times d/2},
%f_i^{gl} = {\rm AvgPool}({\rm GELU}({\rm Linear}({\rm LayerNorm}(x_i))))  \in \mathbb{R}^{d/2}.
%\frac{\sum_{i=1}^{N}t_i*a_i}{\sum_{i=1}^{N}a_i} {\rm MLP}(x_i)  \in \mathbb{R}^{N\times S}
\label{eq:mlp2} 
\end{equation}
where $D$ is the keep/prune decision of the current tokens evaluated by Eq.~\eqref{eq:gumbel}. We then pass the combined feature $f_i {=} [f_i^{local},f_i^{global}] {\in} \mathbb{R}^{N\times d}$ through a MLP pipeline of $Linear(d,d/2)$$\rightarrow$$GELU$$\rightarrow$$Linear(d/2,d/4)$$\rightarrow$$GELU$$\rightarrow$ $Linear(d/4,2)$ to produce a series of token score maps $\{t_i\}^{H}_{i=1}{\in} \mathbb{R}^{N\times 2}$, with $t_i$ indicating the token score from each attention head:
\begin{equation}
\begin{gathered}
%t_i = {\rm Softmax}({\rm MLP}(f_i)) \in \mathbb{R}^{N\times 2}, \\
t_i = {\rm Softmax}({\rm MLP}(f_i)) \in \mathbb{R}^{N\times 2}, \\
\end{gathered}
\end{equation}
where $N{\times} 2$ represents the keep and prune probabilities of $N$ number of tokens.

\begin{table}[t!]
%\vspace{-0.5 cm}
\begin{minipage}{0.48\linewidth}
    \centering
    \captionsetup{type=figure}
    \includegraphics[width=1.0\columnwidth]{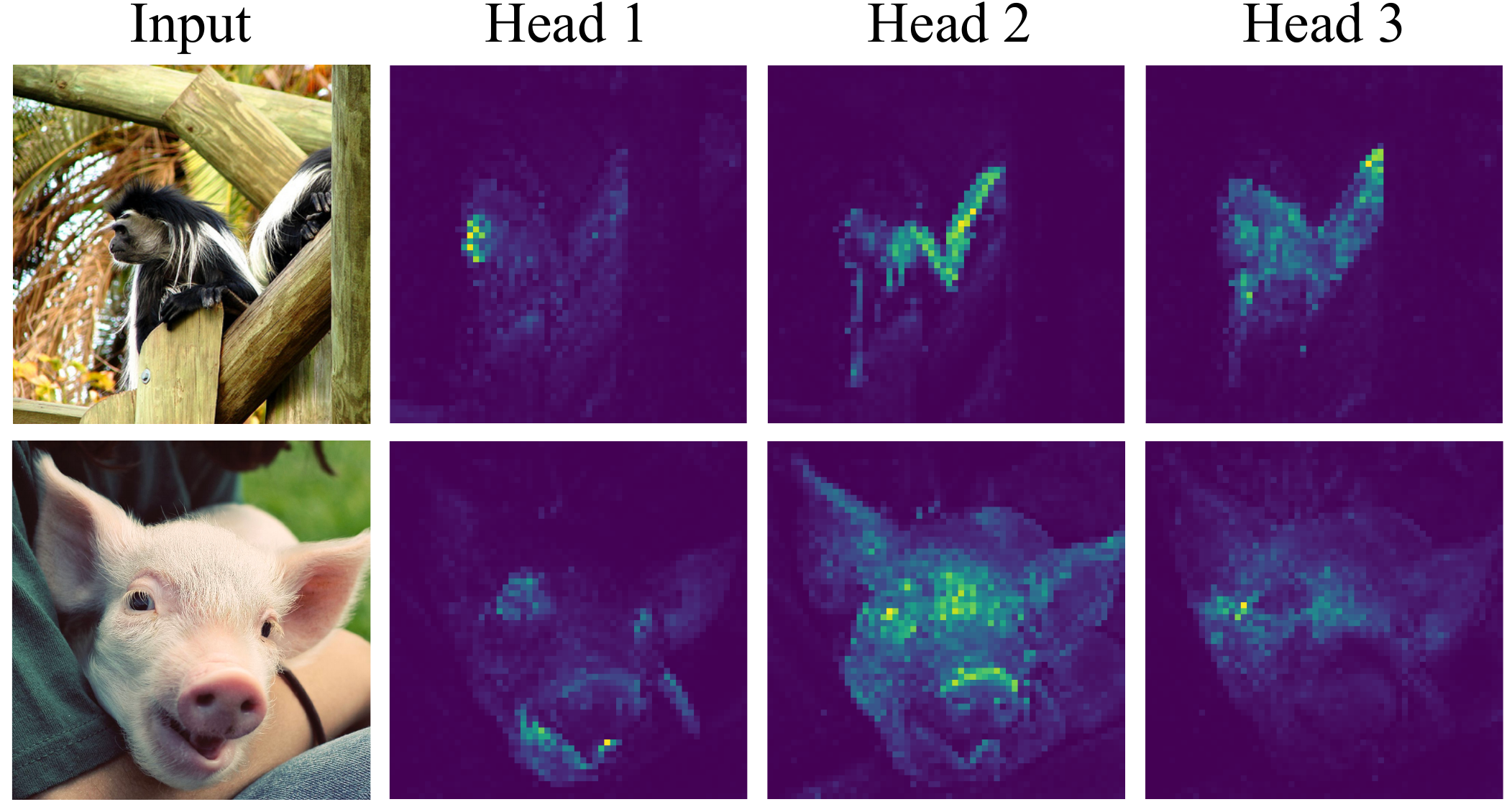}
    \caption{Heatmaps showing the informative region detected by each head in DeiT-T. Each attention head focuses on encoding different image features and visual receptive fields.}
    \label{fig:headattention}
	\end{minipage}\hfill
	\begin{minipage}{0.50\linewidth}
        \centering
        \captionsetup{type=figure}
        \includegraphics[width=1.0\columnwidth]{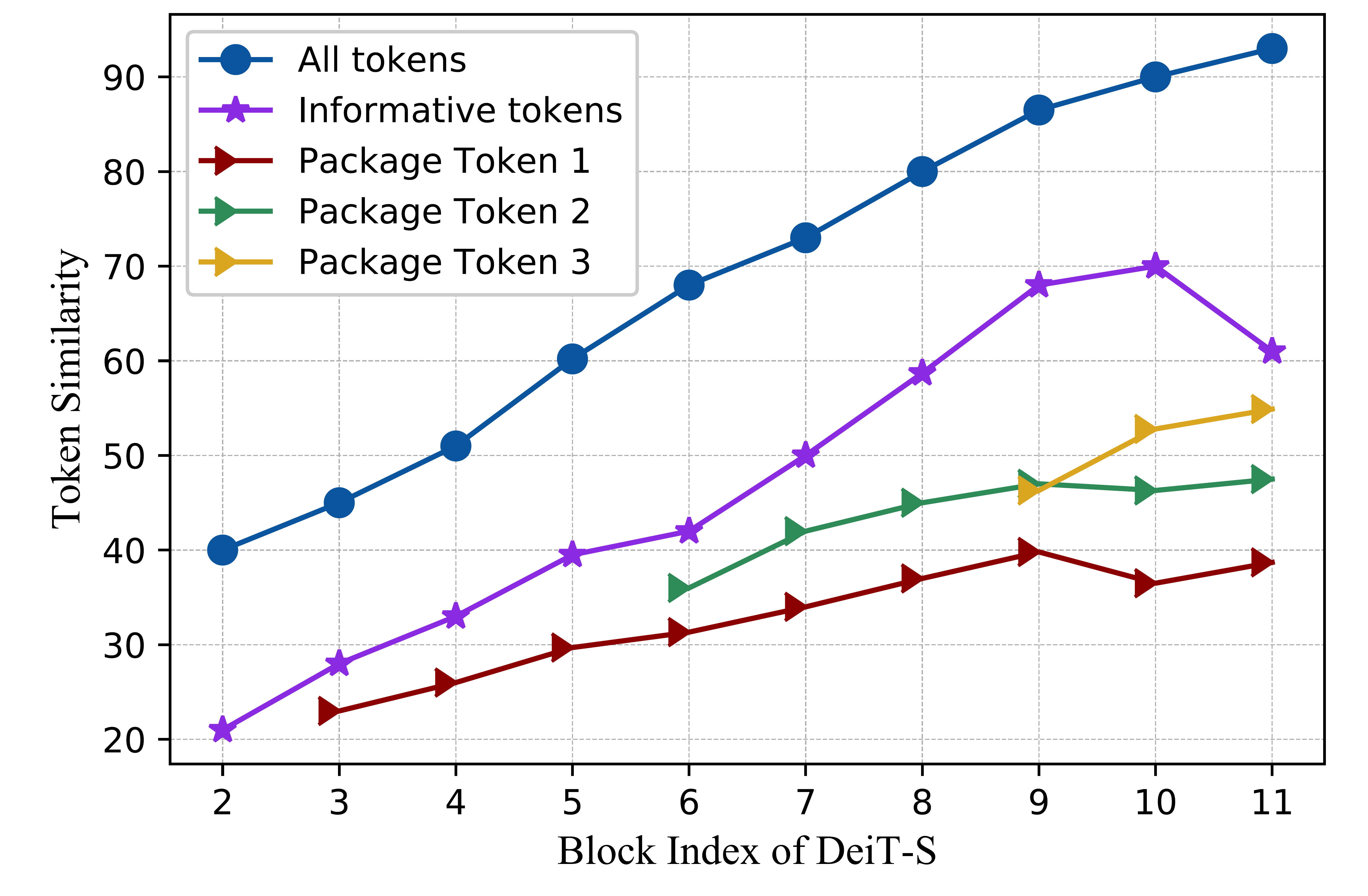}
        \caption{The CKA between the final CLS token and other tokens.}
        \label{fig:token_sim}
	\end{minipage}
	%\caption{The CKA between the final CLS token and other tokens.}
%%%\vspace{-0.4cm}
\end{table}

\noindent\textbf{Head Attention Branch.}
We merge the individual score maps by the weights of each attention head to get the overall token score. As shown in Fig.~\ref{fig:framework}, %We adopt the same structure from SENET~\textcolor{green}{[REF]} and add it along our selector to evaluate the head importance score. 
we add an attention-based branch along the selector backbone to synthesis the importance of each head:
%%%\vspace{-0.4cm}
\begin{equation}
\bar{X} = {\rm AvgPool}(X) =Concat \{ \frac{1}{C} \sum_{i=1}^{C}x_i \}^{H}_{j=1} \in \mathbb{R}^{N\times H},
\label{eq:avg_pool}
\end{equation}
%%%\vspace{-0.2cm}
\begin{equation}
A = {\rm Sigmoid} ({\rm Linear}({\rm GeLU} ({\rm Linear}(\bar{X}))))  \in \mathbb{R}^{N\times H},
\label{eq:sigmoid_fc}
\end{equation}
where $\bar{X}$ is a head-wise statistic generated by shrinking $X$ through its channel dimension $C$ with global average pooling. In Eq.~\eqref{eq:sigmoid_fc}, the attention head score vector $A$  is obtained by feeding $\bar{X}$ into the $ Linear(H,H/2) {\rightarrow} GeLU{\rightarrow} Linear(H/2,H) {\rightarrow} Sigmoid$ pipeline to fully capture head-wise dependencies.
The overall token score is calculated by adding the token scores from each individual  attention head, multiplying by their individual head score $\{ a_i\}^{H}_{i=1}{\in} \mathbb{R}^{N\times 1}$:
%\begin{equation}
%A = [a1,a2,...,ai] \quad 1 \le i \le H 
%\end{equation}
%%%\vspace{-0.2cm}
\begin{equation}
\tilde{T} = \frac{\sum_{i=1}^{H}t_i*a_i}{\sum_{i=1}^{H}a_i}   \in \mathbb{R}^{N\times 2},
\end{equation}
where $\tilde{T}$ is the final token probability score. % importance score of each head.
To make the token removing differentiable, we apply the Gumbel-Softmax technique to generate the token keep/prune decision during training: 
\begin{equation}
D={\rm GumbelSoftmax}(\tilde{T}) \in \{0,1 \}^N.
\label{eq:gumbel}
\end{equation}
%To maintain uniform input token length throughout the model, $D$ serves as a policy to mask the tokens that was determined to be pruned, preventing them from taking part in further computation.
Next, $D$ passes on to the following layers until reaching the next token selector, where it will be updated by applying Hadamard product with the new token keep decision $D\odot D^{\prime}$ during our hierarchical pruning scheme.

Self-attention matrices-based methods~\cite{liang2022evit,xu2022evovit} usually require sorting and evaluating the importance of tokens by a Top-k operation, which is currently not supported in many frameworks for edge devices~\cite{prillo2020softsort}.
On the contrary, our selector generates binary matrices with the help of gumble softmax and FC layers to perform pruning instead of Top-k ordering.
For hardware efficiency, our token selector mainly leverages the FC layers to reuse the GEMM hardware engine already built for the backbone ViT.

\subsection{Token Packaging Technique}
As discussed before, ViT is less accurate for evaluating token values in earlier blocks. Poor scoring may cause important tokens to be removed. Moreover, completely removing background  (negative) tokens will weaken self-attention's ability to capture key information~\cite{yang2021instance}. Instead of completely discarding tokens that are considered less informative, we apply a token packaging technique that integrates them into a package token. Assume there are $Q$ less informative tokens $\hat{X} = \{ n_i\}^{Q}_{i=1}$, $n_i\in \mathbb{R}^{C}$
, along with their token scores $\hat{T} = \{ m_i\}^{Q}_{i=1}$, $m_i\in\mathbb{R}^{2}$
% \begin{equation}
% \begin{gathered}
% \hat{X} = [\hat{x}_1,\hat{x}_2, ..., \hat{x}_l ] \in \mathbb{R}^{L\times C}, \\
% \hat{\tilde{T}} = [\hat{t}_1,\hat{t}_2, ..., \hat{t}_l ] \in \mathbb{R}^{L\times 2}, \quad 1  \le i  \le L .
% \end{gathered}
% \end{equation}
These tokens are combined into one token by:
\begin{equation}
%P = \frac{ \sum_{i=1}^{Q}n_i*m_i}{\sum_{i=1}^{Q}m_i} \in \mathbb{R}^{1\times C},
P = \frac{ \sum_{i=1}^{Q}n_i\cdot m_i[0]}{\sum_{i=1}^{Q}m_i[0]} \in \mathbb{R}^{C},
\end{equation}
where $P$ is the package token; %$m_i$ is an individual token; $n_i$ is its corresponding score. 
$m_i[0]$ is the probability of keeping the token. % $m_i[1]=1-m_i[0]$ is for pruning probability. 
Token $P$ will participate in the subsequent calculations along with the informative tokens, enabling the model to correct scoring mistakes. Our overall framework is efficient, with miniature computation cost (less than 1\% of the total model GFLOPs). All the operations (MLP, Softmax, Pooling, Sigmoid, etc.) are well supported on edge platforms.

\begin{table}[t!]
 \caption{Latency of one DeiT block on the Xilinx ZCU102 FPGA board.}
   \label{tab:latency_ratio}
 \small
 \centering
 %\vspace*{-9pt}
 \resizebox{0.6\columnwidth}{!}{
\begin{tabular}{lcccccc}
\hline\noalign{\smallskip}
 Pruning Rate     &0.0        & 0.1     &0.2  &0.3    &0.4   &0.5 \\ 
 \noalign{\smallskip}
\hline
\noalign{\smallskip}
 DeiT-T Latency (ms)  &0.689     & 0.630 &0.587 &0.509 &0.468 &0.424 \\ 
DeiT-S Latency (ms)   &2.107     &1.891  &1.710 &1.503  &1.315 & 1.121 \\
\bottomrule
\end{tabular}
}
%%%\vspace{-0.4cm}
\end{table}

\subsection{Latency-Aware Training Strategy}
Our latency-aware training strategy includes two parts: (1) the training objective where we introduce the latency-aware sparsity loss to obtain the pruning rate of token constrained by the latency specifications of the target devices; (2) the layer-to-phase progressive training schedule by which we can determine the location of inserted selectors and their suitable pruning rates.

\noindent\textbf{Latency-Sparsity Table.} In order to bridge the inference of ViT model produced by SPViT to the actual latency bound of hardware operation, we measure the latency-sparsity table of the target device, shown in Table~\ref{tab:latency_ratio}. Note that the computation amount of one selector is less than 1\% of one ViT block and the specific latency can be disregarded. More measurement results on edge devices are shown in the Appendix.

\noindent\textbf{Latency-Aware Sparsity Loss.} Based on the relationship between the pruning rate and latency in Table~\ref{tab:latency_ratio}, we introduce a latency-aware sparsity loss $\pounds_{ratio}$:
% \setlength{\belowdisplayskip}{0pt} \setlength{\belowdisplayshortskip}{0pt}
% \setlength{\abovedisplayskip}{0pt} \setlength{\abovedisplayshortskip}{0pt}
%%%\vspace{-0.2cm}
\begin{equation}
%{\rm Block}=Latency-Sparsity(\rho_{i}),
\mathrm{Block\_lat}(\rho_{i})=latency\_sparsity\_table(\rho_{i}),
    \label{eq:single_block}
\end{equation}
%%%\vspace{-0.4cm}
\begin{equation}
    %\sum_{i=1}^{L}{\rm Block}_{i}(\rho_{i}) \leq {\rm LatencyLimit},
    \sum_{i=1}^{L}{\rm Block\_lat}(\rho_{i}) \leq {\rm LatencyLimit},
    \label{eq:hardware_cost}
\end{equation}
%%%\vspace{-0.4cm}
\begin{equation}
    \pounds_{ratio}=\sum_{i=1}^{L}(1-\rho _{i}-\frac{1}{B}\sum_{b=1}^{B}\sum_{j=1}^{N}D_{j}^{i,b})^{2},
    \label{eq:hd_loss}
\end{equation}
where Eq.~\eqref{eq:single_block} is a look-up-table which aims to find the latency of one block $\mathrm{Block\_lat}$ under the corresponding ratio $\rho_i$ with Table~\ref{tab:latency_ratio}.
%${N}'$ is the token number in the current block, and $C$ is the token dimension;
Eq.~\eqref{eq:hardware_cost} guarantees that the inference latency of the model should be under the limit of target edge devices after token pruning. ${LatencyLimit}$ is the latency constraints of the target device. %, which can be estimated based on the hardware resources.
With $i$ being the block index, %$I_{i}$ is a binary variable indicating whether a selector is inserted in $Block_{i}$. 
$\rho_{i}$ is the corresponding pruning rate. % such that ${N}' {=} \rho_{i}{\cdot} N$ is the number of remaining tokens.
Through Eq.~\eqref{eq:single_block} and \eqref{eq:hardware_cost}, we derive appropriate $\rho_{i}$ and feed it to the final sparsity loss (\ref{eq:hd_loss}), where $B$ is the training batch size, and $D_{*}^{i,*}$ (Eq.~\eqref{eq:gumbel}) is token keep decision.
In order to achieve per-image adaptive pruning, we set the average pruning rate of all images in one batch as the convergence target of the Eq.~\eqref{eq:hd_loss}.
% Experiments show that the pruning rate difference of images in the same block will not exceed~4.2\%.

\noindent\textbf{Training Objective.} It includes the standard cross-entropy loss, soft distillation loss, and latency-aware sparsity loss.
The former two are the same as the loss strategy used in DeiT~\cite{Touvron2021TrainingDI}.
\begin{equation}
\pounds = \pounds_{cls}+\lambda _{KL}\pounds_{KL}+\lambda _{distill}\pounds_{distill}+\lambda _{ratio}\pounds_{ratio},
    \label{eq:total_loss}
\end{equation}
where we set $\lambda_{KL}{=}0.5$, $\lambda_{distill}{=}0.5$, $\lambda_{ratio}{=}2$ in all our experiments.

\noindent\textbf{Layer-to-Phase Progressive Training Schedule.} 
Based on~\cite{zhou2021refiner}, we assume that the final CLS token is strongly correlated with classification. And we use centered kernel alignment (CKA) similarity~\cite{kornblith2019similarity} to calculate the similarity of the token features in each block and the final CLS token.
As shown in Fig.~\ref{fig:token_sim}, the final CLS token feature is quite different from token features in earlier blocks. It shows that the representations in earlier blocks are encoded inadequately, which proves the difficulty of pruning tokens in the earlier blocks. Combined with this encoding pattern, we design a latency-aware progressive training strategy to find the optimal accuracy-pruning rate trade-offs and proper locations for token selectors.
In a ViT, tokens can be more effectively encoded in later blocks. Hence, we adopt progressive training on the token selector from later blocks to earlier ones.
Specifically, each time we insert a token selector, we train the current selector and finetune the other parts (backbone and other selectors) by increasing the pruning rate of the current block until accuracy decreases noticeably ($>0.5\%$). We repeat the insertion until there is one selector for each block. Then if the adjacent selectors have a similar pruning rate (difference$<8.5\%$), we combine them as one selection phase and solely keep the first selector of the phase. Finally, if the final computations are lower than the target latency of specific edge devices, we reduce the pruning rate of the first selector. This is because we observe that earlier blocks are more sensitive to pruning.

\section{Experiments}
%\subsection{Datasets and Models}
\noindent\textbf{Datasets and Implementation Details.}
Our experiments are conducted on ImageNet-1K~\cite{5206848} with different backbones including  DeiT-T, DeiT-S~\cite{Touvron2021TrainingDI}; LV-ViT-S, LV-ViT-M~\cite{jiang2021all}; PiT-T, PiT-XS, PiT-S~\cite{heo2021pit}; Swin-T, Swin-S~\cite{liu2021Swin}. The image resolution is 224$\times$224.
We follow most of the training settings as in DeiT and train all backbone models for 60 epochs.
Through our layer-to-phase training, we observe that inserting three token pruning selectors is best for the computation-accuracy tradeoff. 
For DeiT-T/S, we insert the token selector after the 3rd, 6th, and 9th layers. For LV-ViT-S, we insert the token selector after the 4th, 8th, and 12th layers. For LV-ViT-M, we insert the token selector after the 5th, 10th, and 15th layers. For PiT-T/XS/S, we insert the token selector after the 1st, 5th, and 10th layers.
For Swin-T/S, we insert the token selector after each patch merging layer at the 2nd, 3rd, and 4th stage.
%Our searched pruning rate for each phase is 0.383, 0.631, 0.863, respectively.
Our batch size is 256 for DeiT-T, DeiT-S, and LV-ViT-S; and 128 for LV-ViT-M, PiT-T, PiT-XS, and PiT-S. We set an initial learning rate to be 5e-4 for the soft pruning module and 5e-6 for the backbone. The final model has three token selectors. All models are trained on 8 NVIDIA A100-SXM4-40GB GPUs.
The latency is measured on a Samsung Galaxy S20 cell phone that has  Snapdragon 865 processor, which consists of an Octa-core Kryo 585 CPU.

\begin{figure}[t]
\centering
\includegraphics[width=0.8\columnwidth]{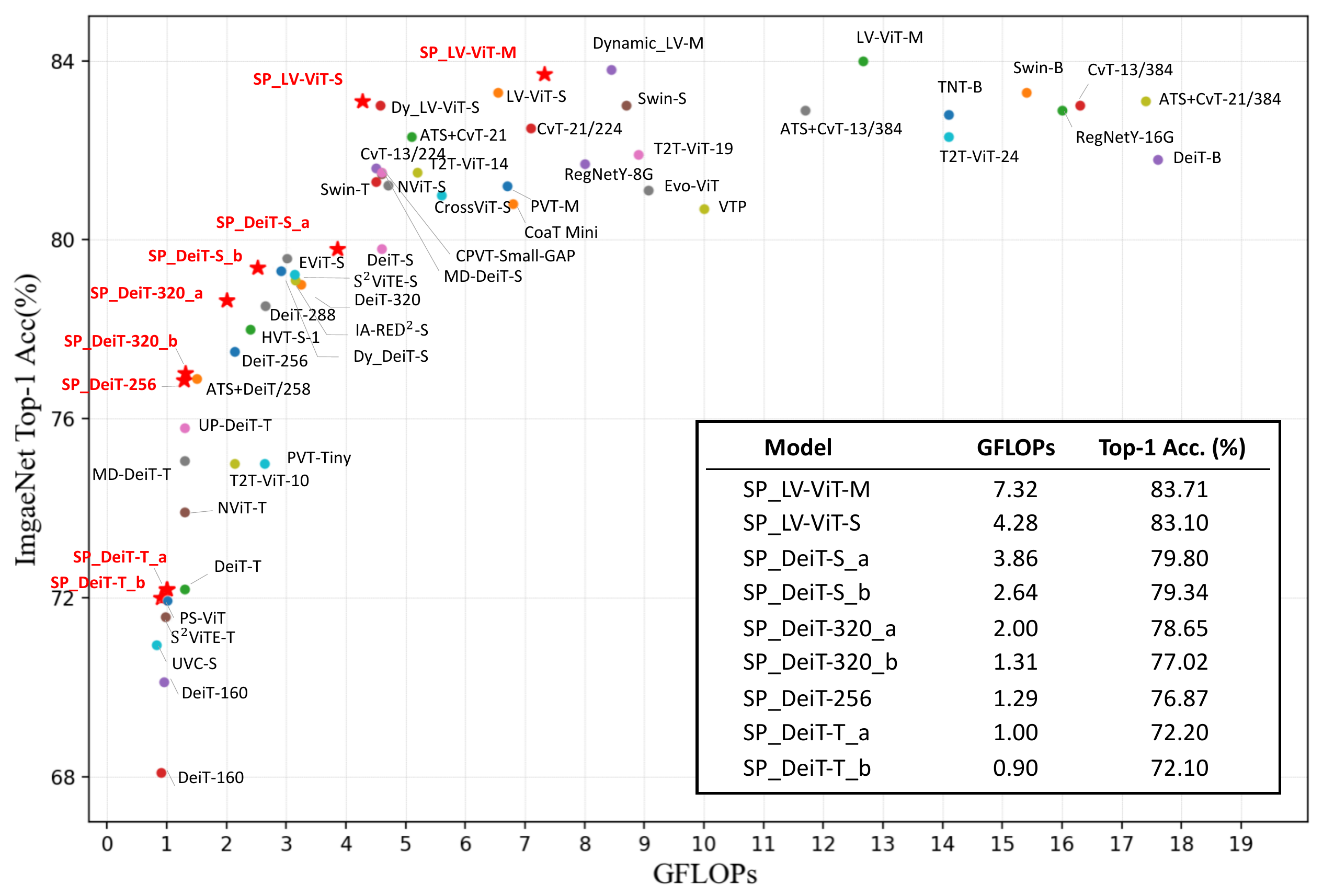}
\caption{Computation (GFLOPs) and top-1 accuracy trade-offs on ImageNet. Our models can achieve better trade-offs compared to other pruned or scaled models.}% GFLOPs and accuracy reductions are calculated from their corresponding dense model.} %Note that ``*'' refers to our reproduced results to obtain models with similar FLOPs for comparison. Baseline/160/192/288/384/512/768 indicates the embedding dimensions.}
\label{fig:main_flops}
%%%\vspace{-0.4cm}
\end{figure}

\subsection{Experimental Results}
\noindent\textbf{Main Results.} We compare our method with several representative methods including DynamicViT~\cite{rao2021dynamicvit}, IA-RED${^2}$~\cite{pan2021iared2}, RegNetY~\cite{radosavovic2020designing}, CrossViT~\cite{chen2021crossvit}, VTP~\cite{zhu2021visual}, ATS~\cite{fayyaz2021}, CvT~\cite{Wu_2021_ICCV}, PVT~\cite{wang2021pyramid}, T2T-ViT~\cite{yuan2021tokens}, UP-DeiT~\cite{yu2021unified}, PS-ViT~\cite{tang2021patch}, Evo-ViT~\cite{xu2022evovit}, TNT~\cite{han2021transformer}, HVT~\cite{pan2021scalable}, Swin~\cite{liu2021Swin}, CoaT~\cite{xu2021co}, CPVT~\cite{chu2021conditional}, EViT~\cite{liang2022evit}, UVC~\cite{yu2022unified},  MD-DeiT~\cite{jia2021efficient},and S${^2}$ViTE~\cite{chen2021chasing}.
%As shown in Table~\ref{main_results}, we report the top-1 accuracy and GFLOPs for each model. Note that ``*'' refers to the results reproduced with similar GFLOPs for comparison. 
Fig.~\ref{fig:main_flops} demonstrates that our models achieve better accuracy-computation trade-offs compared to other pruned or scaled models.
Our SPViT reduces the computation cost by $31\% {\sim} 43\%$ for various backbones with negligible $0.1\% {\sim} 0.5\%$ accuracy degradation, which outperforms existing methods on both accuracy and efficiency.
On lightweight ViT, DeiT-T, the proposed SPViT still reduces GFLOPs by $31\%$ with a negligible $0.1\% $ decrease of accuracy (72.10\% vs. 72.20\%).
To explore model scaling on ViT, we train more DeiT models with the embedding dimension of 160/256/288/320 as our baselines.
On DeiT-T and DeiT-S under the same or similar GFLOPs, the accuracy improvement of SPViT over 
DeiT-160 is 4\% (72.1\% vs. 68.1\% with $\sim$ 0.9 GFLOPs), 4.67\% (76.87\% vs. 72.20\% with $\sim$1.3 GFLOPs) of SPViT-256 over DeiT-T-192, 4.82\% (77.02\% vs. 72.20\% with $\sim$1.3 GFLOPs) of SPViT-320 over DeiT-T-192, and 0.81\% (79.34\% vs. 78.53\% with $\sim$2.65 GFLOPs) of SPViT over DeiT-S-288.
% 4.67\% on DeiT-256 (vs. DeiT-T), 4.82\% on DeiT-320 (vs. DeiT-T), and 0.8\% on DeiT-S (vs. DeiT-288). 
Additionally, our method can prune up to 23.1\% on DeiT-T and 16.1\% on DeiT-S without any accuracy degradation. 
%Detailed results are shown in the Appendix.

% \begin{figure}[t]
% \centering
% \includegraphics[width=0.7\columnwidth]{Figs/main_flops_acc_test_addscaling_nopit4.pdf}
% \caption{Computation (GFLOPs) and top-1 accuracy trade-offs on ImageNet. Our models can achieve better trade-offs compared to other pruned or scaled models.}% GFLOPs and accuracy reductions are calculated from their corresponding dense model.} %Note that ``*'' refers to our reproduced results to obtain models with similar FLOPs for comparison. Baseline/160/192/288/384/512/768 indicates the embedding dimensions.}
% \label{fig:main_flops}
% \vspace{-0.4cm}
% \end{figure}

% \begin{table}[t]
%  \small
%   \centering
%  \resizebox{0.6\columnwidth}{!}{
% \begin{tabular}[width=1.0\linewidth]{@{}llcc@{}}\\
%     \toprule
%     \textbf{Model}    & \textbf{Method} & \textbf{GFLOPs} & \textbf{Top1 Acc (\%)}  \\ 
%     \midrule
% PiT-S                  & Base Model      & 2.90    & 80.90                          \\
% PiT-S                  & \textbf{SPViT (Ours)}     & 2.58     & \textbf{80.90}              \\ \hline
% PiT-XS                 & Base Model       & 1.40     & 78.10             \\
% PiT-S                  & \textbf{SPViT (Ours)}     & 1.42     & \textbf{79.01}       \\ \hline
% PiT-T                  & Base Model         & 0.71     & 73.00          \\
% PiT-XS                 & \textbf{SPViT (Ours)}      & 0.72      & \textbf{74.06}     \\
%     \bottomrule
%     \end{tabular}}
%      \caption{Detail analysis on Pooling-based ViT with SPViT.}
%       \label{tab:pit_results}
% \vspace{-0.4cm}
% \end{table}

\begin{table}[t!]
	\begin{minipage}{0.48\linewidth}
		\caption{Evaluation results on Hierarchical Architectures with SPViT.}
		\label{table_pit}
		\centering
		\resizebox{\textwidth}{!}{%
\begin{tabular}[width=1.0\linewidth]{@{}lcc@{}}\\
    \toprule
    \textbf{Model}     & \textbf{GFLOPs} & \textbf{Top1 Acc (\%)}  \\ 
    \midrule
Swin-S                      & 8.70    & 83.20                          \\
\textbf{SPViT (Ours)}       & 6.35 ($26.4\%\downarrow$)    & \textbf{82.71} ($0.49\%\downarrow$)             \\ \hline
Swin-T                      & 4.50     & 81.20            \\
\textbf{SPViT (Ours)}       & 3.47 ($23.0\%\downarrow$)     & \textbf{80.70} ($0.50\%\downarrow$)      \\ \hline
PiT-S                       & 2.90    & 80.90                          \\
\textbf{SPViT (Ours)}       & 2.22 ($23.3\%\downarrow$)    & \textbf{80.32} ($0.58\%\downarrow$)             \\ \hline
PiT-XS                      & 1.40     & 78.10             \\
\textbf{SPViT (Ours)}       & 1.13 ($18.7\%\downarrow$)    & \textbf{77.86} ($0.24\%\downarrow$)      \\ 
% PiT-Ti                      & 0.71     & 73.0            \\
% \textbf{SPViT (Ours)}       & 0.63 ($10.8\%\downarrow$)    & \textbf{72.74} ($0.26\%\downarrow$)      \\ 
    \bottomrule
    \end{tabular}}
	\end{minipage}\hfill
	\begin{minipage}{0.50\linewidth}
		\caption{Evaluation results on Samsung Galaxy S20 with Snapdragon 865 processor and Xilinx ZCU102 FPGA board.}
		\label{tab:hardware_eva}
		\centering
		\resizebox{\textwidth}{!}{%
		%\small{
\begin{tabular}{llcc}  
%\multirow{2}{*}{\textbf{Model}} & \multirow{2}{*}{\textbf{Method}} & \multicolumn{3}{c}{\textbf{Latency (ms)}} \\
%~ & ~ & Attention & MLP & Total 
\toprule
 \textbf{Model}         & \textbf{Method} & \textbf{Top-1 Acc. (\%)} & \textbf{Latency (ms)} 
\\ \hline
 \multicolumn{4}{c}{Samsung Galaxy S20} \\ \hline
 \multirow{2}{*}{DeiT-T}                    & Baseline       &72.20         & 44                   \\
%\multirow{2}{*}{DeiT-T}      & S$^2$ViTE         &70.12         &  35                  \\
%               & DynamicViT     &71.85    &  37                  \\
                        ~ & \textbf{SPViT (Ours)}  &\textbf{72.10}         & \textbf{26}               \\ \hline
\multirow{2}{*}{DeiT-S}  & Baseline       &79.80         & 113                           \\
%                        & S$^2$ViTE        &79.22         & 78                    \\
%\multirow{2}{*}{DeiT-S}   & IA-RED$^2$       &79.10         & 80                   \\
%                        & DynamicViT   &79.30         &  72                  \\
                        ~ & \textbf{SPViT (Ours)}  &\textbf{79.34}        & \textbf{60}                    \\ \hline
                        
 \multicolumn{4}{c}{Xilinx ZCU102 FPGA} \\ \hline
\multirow{2}{*}{DeiT-T} & Baseline    &72.20        & 8.81 \\
                      ~ & \textbf{SPViT (Ours)}  &\textbf{72.10}       & \textbf{5.60} \\ \hline
\multirow{2}{*}{DeiT-S} & Baseline    &79.80         & 22.31 \\
                      ~ & \textbf{SPViT (Ours)}  &\textbf{79.34}        & \textbf{13.23} \\ 
\bottomrule
%\hline
\end{tabular}}
% 		\caption{Evaluation results on Xilinx ZCU102 FPGA board.}
% 		\label{tab:fpga_eval}
% 		\centering
% 		\resizebox{\textwidth}{!}{%
% 		\small{
% \begin{tabular}{llcc}
% \toprule
% \textbf{Model} & \textbf{Method}& \textbf{Top-1 Acc. (\%)}  &\textbf{Latency (ms)} 
% \\ \hline
% \multirow{2}{*}{DeiT-T} & Baseline    &72.20        & 8.81 \\
%                       ~ & \textbf{SPViT (Ours)}  &\textbf{72.10}       & \textbf{5.60} \\ \hline
% \multirow{2}{*}{DeiT-S} & Baseline    &79.80         & 22.31 \\
%                       ~ & \textbf{SPViT (Ours)}  &\textbf{79.34}        & \textbf{13.23} \\ 
% \bottomrule
% \end{tabular}
% }}
	\end{minipage}
	%%%\vspace{-0.4cm}
\end{table}

\noindent\textbf{Results on Hierarchical Architectures.}
We also perform SPViT on lightweight hierarchical ViTs: Swin-Transformer and PiT, and present the results in Table \ref{table_pit}.
Our SPViT reduces the computation cost by $23\%{\sim} 27\%$ for Swin with a slight accuracy degradation of $0.4\%{\sim}0.5\%$, and by $18\%{\sim}24\%$ for PiT with a degradation of $0.2\%{\sim}0.6\%$.
Even though Swin has scaled down the computation complexity to $O(N)$ through window-based self-attention, and PiT is already a lightweight ViT model,  we still can achieve a fair amount of compression while keeping the accuracy intact.

\subsection{Deployment on Edge Devices}
To evaluate the hardware performance, we implement a framework that runs the ViT model on edge devices. The evaluation is conducted on a Samsung Galaxy S20 cell phone that has a Snapdragon 865 processor, which consists of an Octa-core Kryo 585 CPU carrying high performance with good power efficiency. We use all eight cores on mobile CPUs. We report the average latency of over 100 inferences. %As shown in Table~\ref{tab:hardware_eva}, our method outperforms existing pruning methods on both latency and accuracy. 
As shown in Fig.~\ref{fig:main_latency}, our method outperforms existing pruning methods on both latency and accuracy. 
The deficiencies of other methods mainly lie in three categories: limited pruning capability (low pruning rate)~\cite{pan2021iared2}, non-optimal pruning dimension (number of heads)~\cite{chen2021chasing}, and less efficient operators (e.g., Argsort.)~\cite{rao2021dynamicvit}.
%Fig.~\ref{fig:main_latency} shows a comparison of different models with various accuracy-latency trade-off. 
As shown in Table~\ref{tab:hardware_eva}, on the one hand, our models can outperform lightweight models such as DeiT-T by up to 4.8\% under similar latency. On the other hand, we are able to reduce the latency of larger models such as DeiT-S by up to 47\% (60ms vs. 113ms) with only 0.46\% decrease of accuracy.
Especially, for DeiT-T, we achieve 26 ms per inference on mobile CPUs, which meets the real-time
%\qin{requirement, i.e., 30 inference per seconds} \textcolor{red}{it is better to define what is real-time in the Intro.} 
requirement. As far as we know, this is the first demonstration of ViT inference over 30 fps on edge devices.

\begin{figure*}[t!]
\centering
\includegraphics[width=1\columnwidth]{ 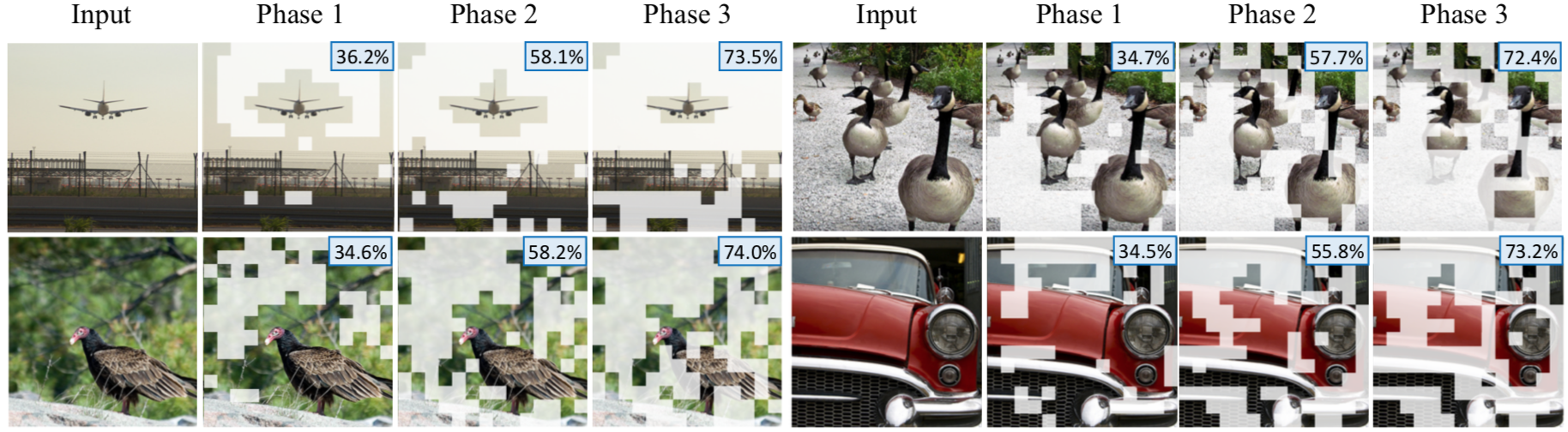}
\caption{Visualization of each pruning phase. In the 1st phase, the selector removes part of the background. In the 2nd phase, it targets the object of interest closely. In the 3rd phase, it localizes the informative features of the objects. The top right corner of each image shows the pruning rate after each phase.}
\label{fig:visual_package_main}
%%%\vspace{-0.4cm}
\end{figure*}

Additionally, SPViT is evaluated on an embedded FPGA platform, Xilinx ZCU102. To maintain the model accuracy on hardware, 16-bit fixed-point precision is adopted to represent all the model parameters and activation data. The comparison results with baseline models are shown in Table~\ref{tab:hardware_eva}. In addition to the total latency, the average latency of the multi-head attention and MLP modules in each model is listed. Compared with the baseline, DeiT-T and DeiT-S, SPViT could achieve 1.57$\times$ and 1.69$\times$ acceleration in the total latency, respectively.

\subsection{Token Pruning Visualization}
%\textbf{Visualizations}
We further visualize the hierarchical token reduction process of SPViT within Fig.~\ref{fig:visual_package_main}. We show the input images along with their sparsification results after each phase. The masked regions represent the tokens that have been soft pruned. Our SPViT can gradually drop less informative tokens and preserve the tokens that contain representative regions with an adaptive pruning rate for each image.

\subsection{Ablation Analysis}

\begin{table}[t!]
	\begin{minipage}{0.45\linewidth}
		\caption{Token selector number/location evaluation on DeiT-S.}
		\label{selector_location}
		\centering
		\resizebox{\textwidth}{!}{%
\begin{tabular}{lccc}
\toprule
\bf Location &\bf Params (M) &\bf GFLOPs &\bf Top-1 Acc. (\%)   \\
 \hline
  3-6-9          &22.13     &\textbf{2.65}      & \textbf{79.34} \\
  1-6-9          &22.13     &2.70      & 76.10 \\
  3-6-11         &22.13     &2.72      &78.76  \\
  6-9            &22.10     &2.71      &78.53  \\
  3-5-7-9        &22.16     &2.66      & 79.34 \\
\bottomrule
\end{tabular}
}%}

% \caption{Package token number comparison on DeiT-S}
% \label{1keep_3keep}
% \resizebox{\textwidth}{!}{
% \begin{tabular}{lccc}
% \toprule
% \bf Method &\bf Params (M) &\bf GFLOPs &\bf Top-1 Acc. (\%)   \\
%  \hline 
% Baseline                        &22.10     &4.60       &79.80    \\
% No Package Token        &22.13     &2.64      &79.21  \\
% 1 $\times$ Package Token        &22.13     &2.64      &79.30  \\
% %\quad  + Attention-based Branch  &22.13     &2.64      &79.30   \\
% 3 $\times$ Package Token          &22.13     &2.64      &\textbf{79.34}   \\
% %\quad  + Attention-based Branch     &22.13     &2.64      &\textbf{79.34}   \\
% \bottomrule
% \end{tabular}
% }
\caption{Comparison of different pruning methods.}
		\label{diff_prune}
		\centering
		\resizebox{\textwidth}{!}{
\begin{tabular}{llcc}
\toprule
 \textbf{Model}   &\bf Method &\bf GFLOPs &\bf Top-1 Acc. (\%)   \\
 \hline 
~&Random                  & 0.90              & 69.87       \\
DeiT-T &Structure              & 0.90              & 70.32       \\
%Attention               & 0.9             &       \\
~&Token selector          & \textbf{0.90}    & \textbf{72.10}      \\
\hline

~&Random                  & 2.64            & 77.25       \\
DeiT-S &Structure          & 2.64              & 77.86       \\
%Attention               & 0.9             &       \\
~&Token selector          & \textbf{2.64}    & \textbf{79.34}      \\
\bottomrule
\end{tabular}}
	\end{minipage}\hfill
	\begin{minipage}{0.51\linewidth}

        \centering
        \captionsetup{type=figure}
        \includegraphics[width=1.0\columnwidth]{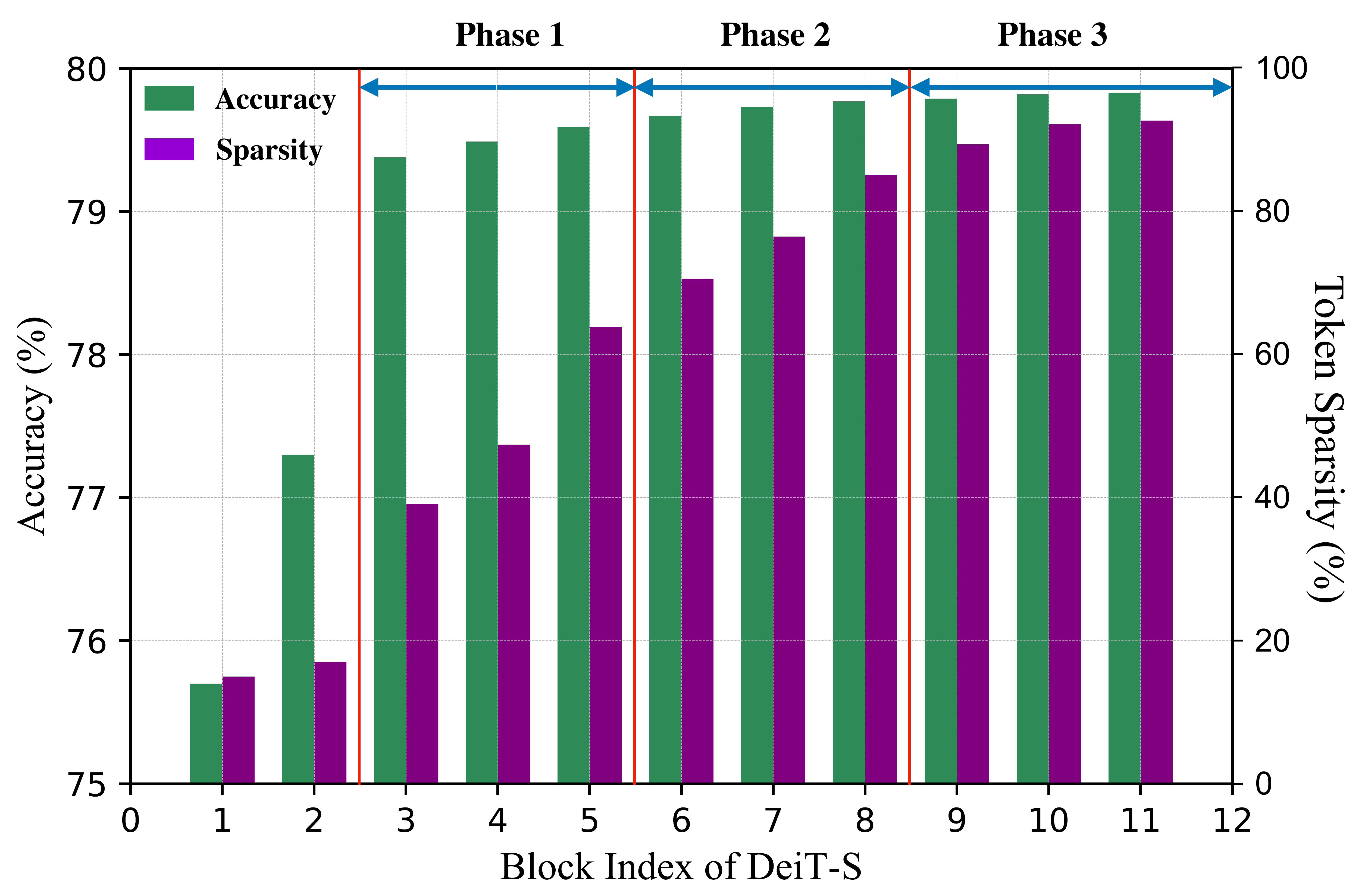}
        \caption{The accuracy and the token sparsity distribution after the Layer-to-Phase Progressive Training. We do the insertion behind the $Block_{index}$. Our final phase plan is demonstrated above.}
        \label{fig:model_location}
	\end{minipage}
%%%\vspace{-0.4cm}
\end{table}

\noindent\textbf{Token Selector Number and Location.} After progressive training (each selector is fine-tuned by 25 epochs), we can get the pruning rate of each block as shown in Fig.~\ref{fig:model_location}.
Based on the trend of the figure, we can divide the evolution of the pruning rate into 2 phases, 3 phases, and 4 phases.
We keep the appropriate selectors accordingly and re-finetuning the whole model.
In Table~\ref{selector_location}, the 3-6-9 division style has the highest accuracy and the lowest computation cost, just like 3-5-7-9. According to the test on Samsung Galaxy S20, each selector and corresponding package token will introduce a delay of 1.67 ms, so we choose 3-6-9 as the best.
For another 3-phase style, 1-6-9, the accuracy and computation cost are both not ideal.
This shows that due to insufficient encoding, it is difficult to perform token pruning in the earlier blocks of ViTs.
Meanwhile, for the 3-6-11 style, both the accuracy and computation cost are slightly inferior to the 3-6-9 style.
The possible reason is the pruning rate of the second phase should be smaller than the third phase and the coverage of the second phase is too wide.
As a result, there is still a lot of redundancy in the tokens of the third phase, restricting the accuracy and computation efficiency of the model at the same time.
Furthermore, because of a similar reason, the 2-phase style, 3-6, cannot achieve a better trade-off between accuracy and the computation cost. %More details can be seen in the Appendix.

% \begin{figure}[t]
% \centering
% \includegraphics[width=0.7\columnwidth]{Figs/per_layer.pdf}
% \caption{The accuracy and the token sparsity distribution after the Layer-to-Phase Progressive Training. We do the insertion behind the $Block_{index}$. Our final phase plan is demonstrated above.}
% \label{fig:model_location}
% \vspace{-0.4cm}
% \end{figure}

\noindent\textbf{Comparison of Different Pruning Methods}
To further prove the effectiveness of our score-based dynamic token pruning method, we compare with some general pruning methods: random pruning and structure pruning.  For random pruning, we randomly remove the input token, neglecting the token importance. For structure pruning, we prune the input feature map by dimension, which will impair every token. Results are shown in Table~\ref{diff_prune}. Under the same computational complexities (0.9 GFLOPs for DeiT-T and 2.64 GFLOPs for DeiT-S), our proposed method achieves the best accuracy.

\subsection{Limitations}
For the algorithm design, it might be more effective to combine our framework with the weight pruning strategy for larger ViTs. For the hardware deployment, large amounts of data movement bring much pressure to the memory due to multiple blocks and many intermediate results, which will be optimized in our further work.

\section{Conclusion}
In this paper, we propose a dynamic, latency-aware soft token pruning framework called SPViT. Our attention-based multi-head token selector and token packaging technique, along with the latency-aware training strategy can well balance the tradeoff between accuracy and specific hardware constraints. We deploy our model on mobile and FPGA, which both meet the real-time requirement.

\noindent\textbf{Acknowledgments.}
The research reported here was funded in whole or in part by the Army Research Office/Army Research Laboratory via grant W911-NF-20-1-0167 to Northeastern University. Any errors and opinions are not those of the Army Research Office or Department of Defense and are attributable solely to the author(s). This research is also partially supported by National Science Foundation CCF-1919117 and CMMI-2125326.

%\clearpage\mbox{}Page \thepage\ of the manuscript.
%\clearpage\mbox{}Page \thepage\ of the manuscript.

% This is the last page of the manuscript.
% Now we have reached the maximum size of the ECCV 2022 submission (excluding references).
% References should start immediately after the main text, but can continue on p.15 if needed.

\clearpage
\bibliographystyle{splncs04}
\bibliography{egbib}
\newpage
\appendix

\section{Analysis and Discussion}

\subsection{Model Scaling} 
\begin{figure}[htb]
\centering
\includegraphics[width=0.7\columnwidth]{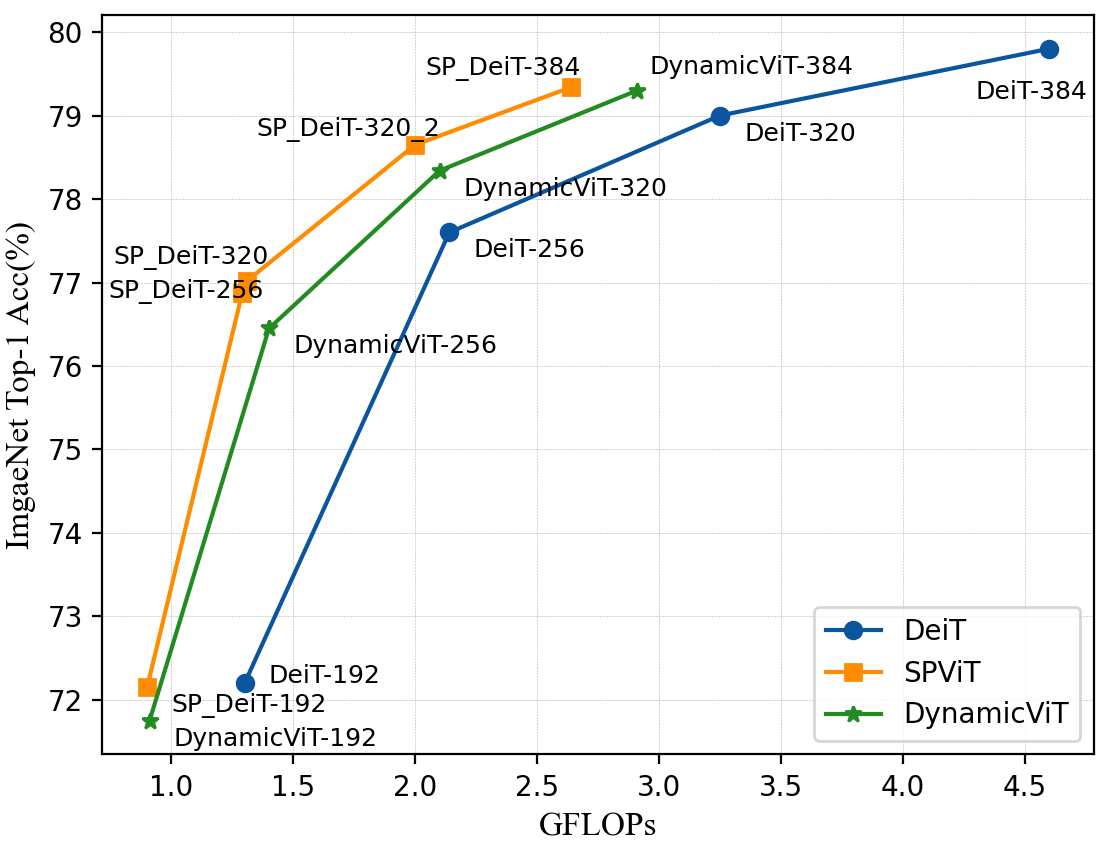}
\caption{Comparison of our SPViT method with model scaling. We prune DeiT models with embedding dimensions varying from 192 to 384 and compare with DynamicViT under comparable GFLOPS.}
\label{fig:model_scaling}
\vspace{-0.4cm}
\end{figure}

In ViTs, the most common method to scale the model is to change the number of channels, while our SPViT provides another perspective to perform token pruning for better complexity/accuracy trade-offs. We illustrate this superior effect of SPViT in Figure~\ref{fig:model_scaling}. 
First, we train several DeiT [25] models with varying embedding dimensions from 192 to 384. 
Second, we compress these DeiTs into ones whose channel has one less head than them.
However, these new compressed models (SPViT-serious) have better accuracy than DeiT original variants with similar computation.
Specifically, the orange line of SPViT-serious is closer to the upper left corner of the Figure~\ref{fig:model_scaling} than the original DeiT-serious.
For DynamicViT, we have also observed uniform benefits, but its efficient models are still slightly inferior to SPViT.
%We also compare SPViT with the model scaling on the PiT.When compressing PiT-S to the comparable size of PiT-XS, the accuracy of the produced model is 0.91\% higher than the original PiT-XS; When the target size is PiT-T, the accuracy of the produced model is 0.9\% higher. %\cross{than the original PiT-T}.

\subsection{Progressive Training Sparsity for Each Layer}
As a supplement for Figure~\ref{fig:model_location}, we show the exact sparsity and accuracy of each layer for our progressive training in Table~\ref{each_layer}. We start by adding the token selector one by one from the 11th layer to the 1st layer. The layer index indicates the layer before the token selector. Between 6$\sim$8 and 9$\sim$11 layer, each layer has similar accuracy and sparsity, which indicates that these layers can be combined to one pruning phase with a token selector at the front. 
\begin{table}[htb]
\begin{center}
\resizebox{0.75\columnwidth}{!}{
\begin{tabular}{lccccccccccc}
\toprule
\bf Layer & 1 & 2 & 3 & 4 & 5 &6&7&8&9&10&11   \\
 \hline 
Accuracy   & 75.7 & 77.3 &79.4 &79.5 &79.6 &79.7 &79.7 &79.7 &79.8 &79.8 &79.8      \\
Sparsity  & 0.15 & 0.17 &0.39 &0.47 &0.64 & 0.71 &0.77 &0.85 &0.89 &0.92 &0.93   \\
\bottomrule
\end{tabular}
}
\caption{Progressive training sparsity for each layer}
\label{each_layer}
\end{center}
\end{table}

% \begin{table}[htb]
%  \small
%  \caption{Detail analysis on Pooling-based ViT with SPViT.}
%  \label{tab:fpga_eval}
%  \centering
%  \resizebox{0.9\columnwidth}{!}{
% \begin{tabular}[width=\linewidth]{@{}llll@{}}\\
%     \toprule
%     \textbf{Model}         & \textbf{Method} & \textbf{Top1 Acc (\%)} & \textbf{FLOPs (G)}  \\ 
%     \midrule
% PiT-S                  & Base Model        & 80.9                   & 2.90             \\
% % PiT-S                  & Random          & 80.6                   & 2.58            \\ 
% PiT-S                  & \textbf{SPViT (Ours)}     & 80.9          & 2.58            \\ \hline
% PiT-XS                 & Base Model        & 78.10                  & 1.40             \\
% PiT-S                  & \textbf{SPViT (Ours)}     & 79.01         & 1.42            \\ \hline
% % DeiT-T                 & Baseline        & 72.20                  & 1.30             \\
% % DeiT-T                 & SPViT (Ours)     & 72.10                  & 0.91             \\
% % PiT-XS                 & SPViT (Ours)     & \textbf{75.01}         & \textbf{0.91}             \\ \hline
% PiT-T                  & Base Model        & 73.00                  & 0.71            \\
% PiT-XS                 & \textbf{SPViT (Ours)}     & 74.06         & 0.72            \\
% % PiT-T                  & \textbf{SPViT (Ours)}     & 72.81                  & 0.66            \\ 

%     \bottomrule
%     \end{tabular}}
% \end{table}

\subsection{Model Latency on Hardware}
We show all model latency results tested on Samsung Galaxy S20 in Table ~\ref{tab:latency_hardware}. On the one hand, our models can outperform lightweight models such as DeiT-T by up to 4.8\% with even smaller latency (38ms vs. 44ms). On the other hand, we are able to reduce the latency of larger models such as DeiT-S by up to 47\% (60ms vs. 113ms) with only 0.46\% decrease in accuracy. On LV-ViT-S/M, our SPViT models show better performance, outperforming DynamicViT on both latency and accuracy.

\begin{table}[h!]
 \small
 \centering
 \resizebox{0.75 \columnwidth}{!}{
\begin{tabular}{llcc}  
%\multirow{2}{*}{\textbf{Model}} & \multirow{2}{*}{\textbf{Method}} & \multicolumn{3}{c}{\textbf{Latency (ms)}} \\
%~ & ~ & Attention & MLP & Total 
\toprule
 \textbf{Model}         & \textbf{Method} & \textbf{Top-1 Acc. (\%)} & \textbf{Latency (ms)} 
\\ \hline
% \\
                            & Baseline       &72.20         & 44                   \\
                             & S$^2$ViTE         &70.12         &  35                  \\
                            & DynamicViT     &71.85    &  37                  \\
\multirow{2}{*}{DeiT-T}     ~ & \textbf{SPViT (Ours)}  &\textbf{72.20}         & \textbf{33}  \\ 
                          ~ & \textbf{SPViT (Ours)}  &\textbf{72.10}         & \textbf{26}  \\ 
                         ~ & \textbf{SPViT-256 (Ours)}  &\textbf{76.87}         & \textbf{36}  \\ 
                         ~ & \textbf{SPViT-320 (Ours)}  &\textbf{77.02}         & \textbf{38}  \\ 
\hline
                         & Baseline       &79.80         & 113                           \\
                         & S$^2$ViTE        &79.22         & 78                    \\
                             & IA-RED$^2$       &79.10         & 80                   \\
\multirow{2}{*}{DeiT-S}      & DynamicViT   &79.30         &  72                  \\
                        ~ & \textbf{SPViT-320 (Ours)}  &\textbf{78.65}        & \textbf{47} \\ 
                        ~ & \textbf{SPViT (Ours)}  &\textbf{79.34}        & \textbf{60} \\ 
                        ~ & \textbf{SPViT (Ours)}  &\textbf{77.02}        & \textbf{38} \\ 
\hline
                              & Baseline       &83.30         & 148               \\
\multirow{2}{*}{LV-ViT-S}     & DynamicViT   &83.00         &  114                  \\
                            ~ & \textbf{SPViT (Ours)}  &\textbf{83.10}        & \textbf{89}                    \\ \hline   
                            & Baseline       &84.00         & 269              \\
\multirow{2}{*}{LV-ViT-M}   & DynamicViT   &83.61         &  195                  \\
                          ~ & \textbf{SPViT (Ours)}  &\textbf{73.71}        & \textbf{152}                    \\ %\hline                        
\bottomrule
%\hline
\end{tabular}
 }
  \caption{Evaluation results on Samsung Galaxy S20 with a Snapdragon 865 processor.}
   \label{tab:latency_hardware}
\vspace{-0.4cm}
\end{table}

\subsection{Number of Package Tokens.}
We insert three soft pruning modules for hierarchical pruning for all models. In each module, a new package token is generated. We conduct two ways to pass on the package token for subsequent layers: 
1) Merge the package token generated from the current module with the existing one from the last module by element-wise addition. Therefore, only 1 additional token is added to the input sequence in total. 
2) Concatenate the new package token to the existing ones. Making it 3 additional tokens in total.
Table \ref{1keep_3keep} shows a comparison of the two methods. %Multiple package tokens perform better.
\begin{table}[t]
\centering
\resizebox{0.7\columnwidth}{!}{
\begin{tabular}{lccc}
\toprule
\bf Method &\bf Params (M) &\bf GFLOPs &\bf Top-1 Acc. (\%)   \\
 \hline 
Baseline                        &22.10     &4.60       &79.80    \\
1 $\times$ Package Token        &22.13     &2.63      &79.26   \\
\quad  + Attention-based Branch  &22.13     &2.64      &79.30   \\
3 $\times$ Package Token         &22.13     &2.64      &79.28   \\
\quad  + Attention-based Branch     &22.13     &2.64      &\textbf{79.34}   \\
\bottomrule
\end{tabular}
}
\caption{Package token number comparison on DeiT-S}
\label{1keep_3keep}
%%%\vspace{-0.4cm}
\end{table}

\subsection{Comparison of Different Token Selector Designs}
We analyze the performance of our token selector design by replacing it with different operations. More specifically, we replace the original MLP based pipeline in Eq.~\eqref{eq:mlp1} and~\eqref{eq:mlp2} with a convolution-based pipeline: $Conv1d \rightarrow BatchNorm1d \rightarrow GELU$. We also evaluated different activation functions: RELU and Hardswish. We compare these variants of the token selector on DeiT-T with a complexity of 0.9 GFLOPs. As shown in Table~\ref{selector_design}, under the same training settings, MLP based token selectors outperform convolution-based token selectors (72.10\% vs 71.56\%). Furthermore, GELU function outperforms Hardswish and RELU (71.56\% vs. 71.48\% vs. 71.13\% on Conv1d+3kernel). However, we can not hastily conclude that MLP and GELU are superior to convolution and other activation functions. This may be due to different training difficulties. Hardswish is harder to converge, so larger epochs may be necessary~\cite{Graham_2021_ICCV}. Additionally, Conv1d with 1kernel can mimic MLP’s fully connected layer. The accuracy gap between these two may due to distinct desirable initial learning rates and schedulers. Result also shows that a larger kernel size may be preferable (71.34\% vs. 71.56\% on Conv1d), which can help learn more local representation. We will further invest in these designs in our future work. More specifically, train each design under different training settings and also combine the advantages of MLP and convolution layers.

\begin{table}[htb]
\begin{center}
\resizebox{0.75\columnwidth}{!}{
\begin{tabular}{lcc}
\toprule
\bf Token Selector &\bf GFLOPs &\bf Top-1 Acc. (\%)   \\
 \hline 
MLP+GELU                     &0.90     &\textbf{72.10}       \\
MLP+Hardswish                &0.90     &71.94        \\
Conv1d+3kernel+RELU          &0.90     &71.13       \\
Conv1d+3kernel+GELU          &0.90     &71.56        \\
Conv1d+3kernel+Hardswish     &0.90     &71.48         \\
Conv1d+1kernel+GELU          &0.90     &71.34        \\
\bottomrule
\end{tabular}
}
\caption{Comparison of different token selector layers and activation functions}
\label{selector_design}
\end{center}
\end{table}

\subsection{Effectiveness of the Token Packaging Technique w/o Class Token}
In many recently proposed vision transformer models, the class token was removed and replaced by doing average pooling on the last output feature to aggregate representation from all patch tokens~\cite{Graham_2021_ICCV,zhai2021scaling}.
This process is similar to our token packaging technique, where we also apply average pooling on the removed tokens. We raise a conjecture: Our token packaging technique can be more effective on models that do not rely on the class token. To test our hypothesis, we run a simple experiment by first pre-training a DeiT-T without a class token, and then applying our method both with and without the token packaging technique. As shown in Table~\ref{cls_token}, when training on a DeiT model that contains a class token, our token packaging technique can improve the performance by 0.12\% (72.1\% vs. 72.0\%). When training on a DeiT model without a class token, our token packaging technique can improve the performance by 0.23\% (70.75\% vs. 70.52\%), which verified our assumption.

\subsection{Comparison of Different Batch Sizes}
We run our SPViT on DeiT-S with different batch sizes for ablation. Results in Table~\ref{batch_size} show that accuracy has a slight boost when batch size increases. 

\subsection{Encoding Redundancy of the Pooling Layer}
We make further discussion on Pooling-based ViT (PiT Series). Figure~\ref{figure_pit} shows that the attention matrix before and after SPViT retains great similarity, which enlightens that the encoding redundancy of the pooling-layer mechanism can be recognized precisely by SPViT.

\begin{table}[t!]
	\begin{minipage}{0.52\linewidth}
		\caption{Effectiveness of the token packaging technique w/o class token}
		\label{cls_token}
		\centering
\resizebox{1.0\columnwidth}{!}{
\begin{tabular}{lcc}
\toprule
\bf Model &\bf GFLOPs &\bf Top-1 Acc. (\%)   \\
 \hline 
DeiT-T              & 1.30              & 72.20       \\
SPViT               & 0.90              & 72.10       \\
SPViT w/o package token  & 0.90              & 71.98       \\
\hline
DeiT-T w/o cls token      & 1.29             & 71.42      \\
SPViT                     & 0.87             & 70.75        \\
SPViT w/o package token   & 0.87              & 70.52       \\
\bottomrule
\end{tabular}
}%}

\caption{Different batch size comparison on DeiT-S}
\label{batch_size}
\resizebox{1.0\columnwidth}{!}{
\begin{tabular}{lccc}
\toprule
\bf Batch Size &\bf GFLOPs &\bf Top-1 Acc. (\%) &\bf Top-5 Acc. (\%)   \\
 \hline 
96           &2.65      &79.31  &94.64  \\
128           &2.65      &79.32  &94.64  \\
256           &2.64      &79.34  &94.67  \\
\bottomrule
\end{tabular}
}
	\end{minipage}\hfill
	\begin{minipage}{0.45\linewidth}

        \centering
        \captionsetup{type=figure}
        \includegraphics[width=1.0\textwidth]{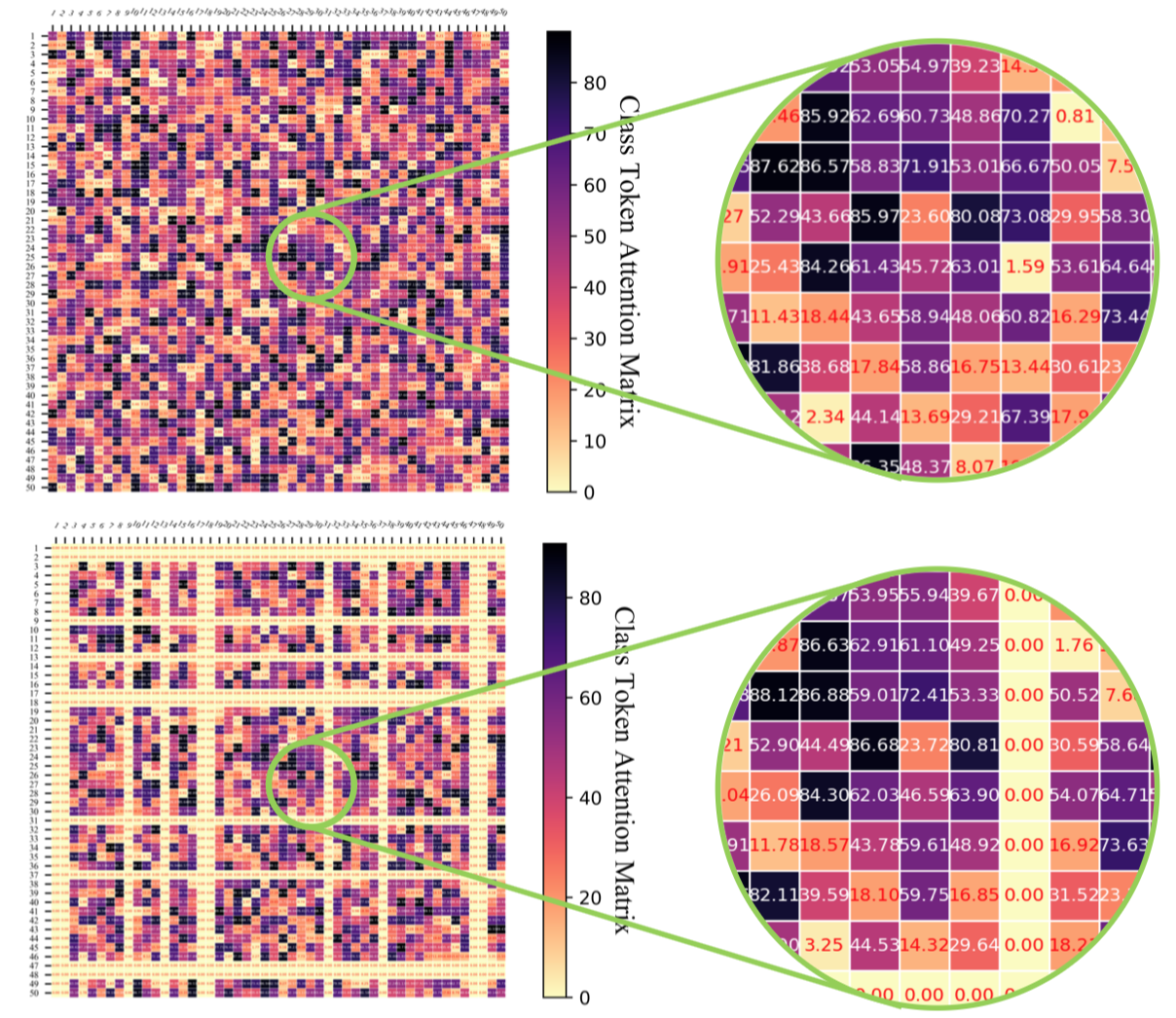}
        \caption{Illustration of the first attention matrix at the final block. Upper figure is the original PiT-S, lower one is with SPViT.}
        \label{figure_pit}
	\end{minipage}
\vspace{-0.4cm}
\end{table}

\section{Visualization}
\subsection{Token Pruning Visualization}
\begin{figure*}[]
\centering
\includegraphics[width=1.0\columnwidth]{  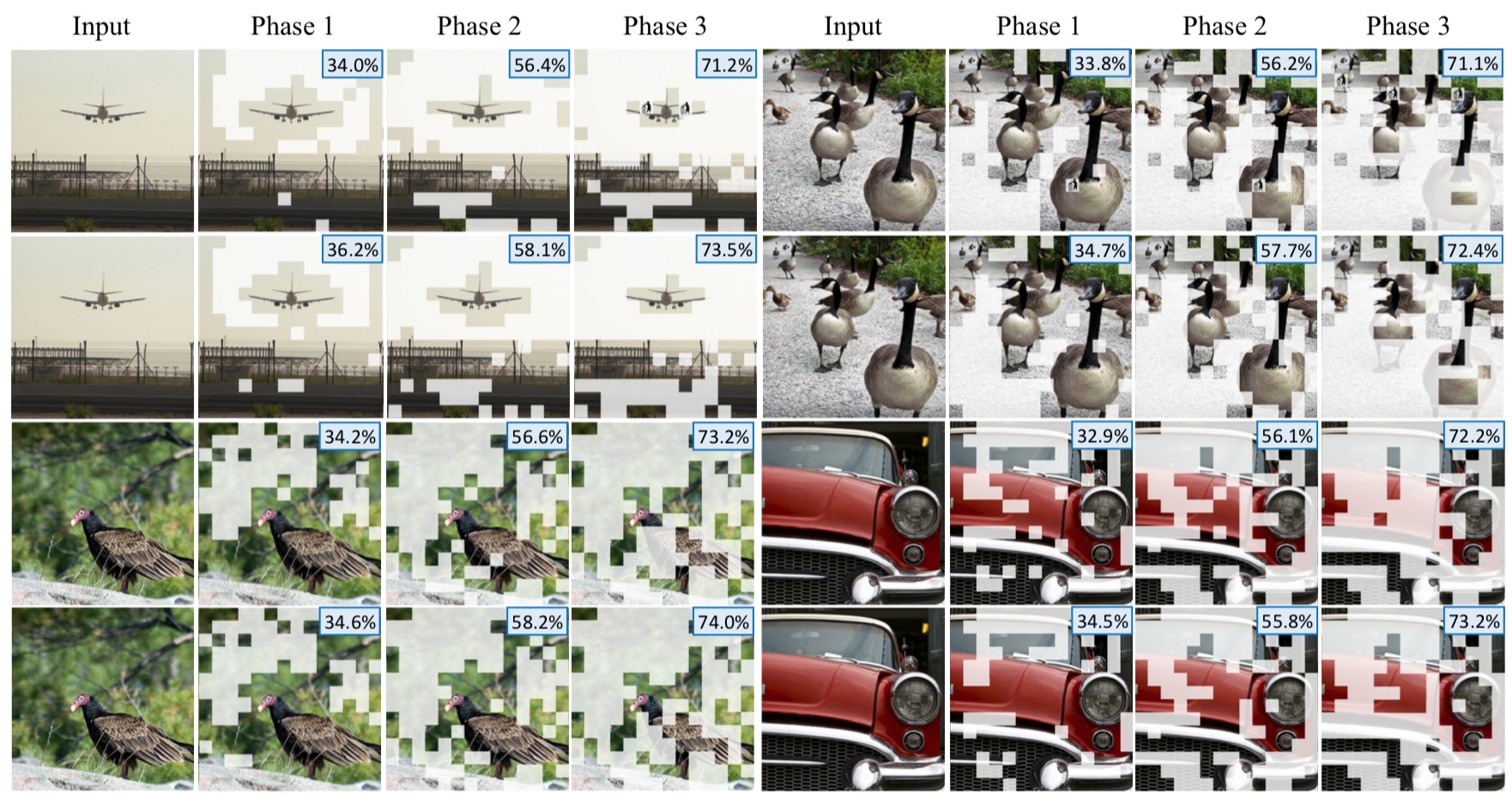}
\caption{Visualization of each pruning phase. Row 1 and 3 show the results collected from the framework without token packaging and row 2 and 4 show the results from the framework with token packaging.}
\label{fig:visual_package}
\vspace{-0.4cm}
\end{figure*}

%\textbf{Visualizations}
We further visualize the process of SPViT to describe the performance in the inference phase and make a comparison between the framework with the token packaging technique and without it the token packaging technique. As shown in Figure~\ref{fig:visual_package}, row 1 and 3 show the results collected from the framework without token packaging, row 2 and 4 are from the framework with token packaging. Take row 1 and 2 for example:
At phase 1, there is a $ 7\%$ difference in the left image groups %between the framework with the token packaging and the w/o token packaging 
and  $11\%$ in the right ones;
At phase 2, $11\%$ difference in the left image groups and $15\%$ in the right ones;
At phase 3, $14\%$ difference in the left image groups and $18\%$ in the right ones.
We can infer token packaging can help to lock the object instead of the background. 
And this phenomenon is more obvious in complex and multi-object images.

\subsection{Self-attention Head Heatmap}
\begin{figure*}[]
\centering
\includegraphics[width=1.0\columnwidth]{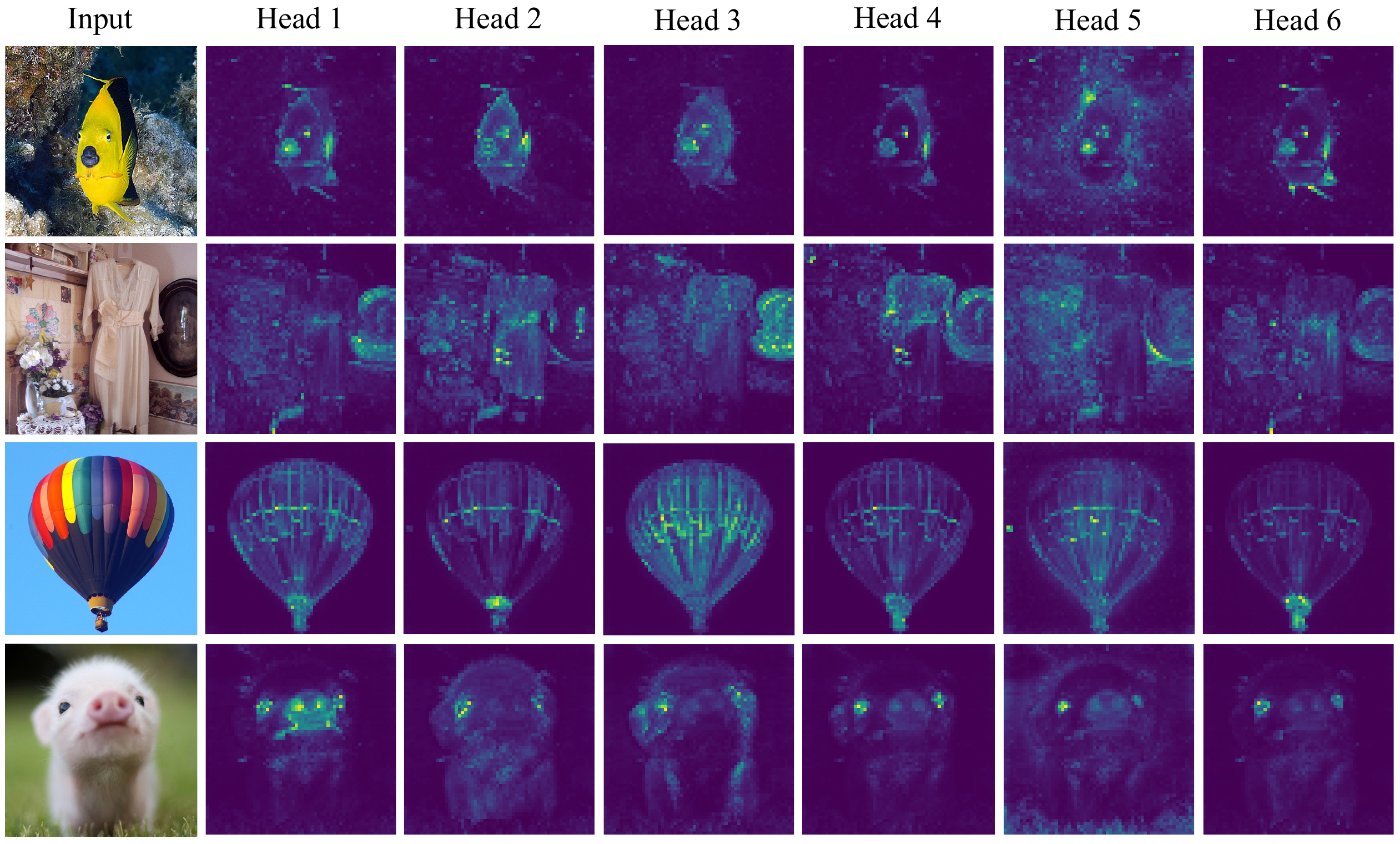}
\caption{Heatmaps showing the informative region detected by each head in DeiT-S. }
\label{fig:headattention_abs}
%\vspace{-0.4cm}
\end{figure*}

Figure~\ref{fig:headattention_abs} shows the heatmaps of informative region detected by each self-attention head in DeiT-S. Each attention head focuses on encoding different image features and visual receptive fields. Therefore, token importance is different for each head. This demonstrates the need for obtaining token score individually.

%\newpage

\section{Main Results}
We compare our method with several representative methods including DynamicViT~\cite{rao2021dynamicvit}, IA-RED${^2}$~\cite{pan2021iared2}, RegNetY~\cite{radosavovic2020designing}, CrossViT~\cite{chen2021crossvit}, VTP~\cite{zhu2021visual}, ATS~\cite{fayyaz2021}, CvT~\cite{Wu_2021_ICCV}, PVT~\cite{wang2021pyramid}, T2T-ViT~\cite{yuan2021tokens}, UP-DeiT~\cite{yu2021unified}, PS-ViT~\cite{tang2021patch}, Evo-ViT~\cite{xu2022evovit}, TNT~\cite{han2021transformer}, HVT~\cite{pan2021scalable}, Swin~\cite{liu2021Swin}, CoaT~\cite{xu2021co}, CPVT~\cite{chu2021conditional}, EViT~\cite{liang2022evit}, UVC~\cite{yu2022unified},  MD-DeiT~\cite{jia2021efficient},and S${^2}$ViTE~\cite{chen2021chasing}.
As shown in Table~\ref{main_results}, we report the top-1 accuracy and GFLOPs for each model. Note that ``*'' refers to the results reproduced with similar GFLOPs for comparison. 
\begin{table*}[t]
\small
\centering
 \resizebox{0.7\columnwidth}{!}{
\begin{tabular}{llcccc}
\toprule
\bf Model  &\bf Method &\bf Params (M)  &\bf GFLOPs &\bf GFLOPs $\downarrow$ (\% )
&\bf Top-1 Acc. (\%) %&\multicolumn{1}{c}{\bf Top-5 Acc. (\%)}  
\\
\hline%{5mm}
                 &Baseline/192   &5.60         &1.30                &0.00             &72.20                \\
                 &Baseline/160*  &4.00         &0.90                &30.77        &68.10                \\
                 &NViT-T         &6.40         &1.30                &0.00             &73.91          \\
                 &UP-DeiT-T      &5.70         &1.30                &0.00             &75.79           \\
                 &MD-DeiT-T      &5.70         &1.30                &0.00            &75.06    \\
                 &T2T-ViT-10     &5.90         &2.13                &-63.85        &75.00           \\
                 &UVC          & -          & 0.64                & -50.74     &  71.3 \\
 DeiT-T                  &PVT-Tiny       &13.20        &2.64                &-103.08            &75.00           \\
                 &DynamicViT     &5.90         &0.91                &30.00        &71.85            \\
                 &PS-ViT         &5.60         &0.93                &28.46        &72.00                 \\
                 &S$^2$ViTE      &4.20         &0.95                &26.92        &70.12                 \\
         &ATS+DeiT/258   &10.13        &1.50                &-15.38       &76.90            \\
                 &\textbf{SPViT (Ours)}        &5.70                &1.00         &23.08         &\textbf{72.20}                \\
                 &\textbf{SPViT (Ours)}        &5.70                &0.90         &30.77         &\textbf{72.10}                \\
                 &\textbf{SPViT-256 (Ours)}   &10.16                &1.29         &0.77         &76.87                \\
                 &\textbf{SPViT-320 (Ours)}   &15.44                &1.30         &0.00         &\textbf{77.02}                \\
 \hline 
                 &Baseline/384   &22.10       &4.60                 &0.00              &79.80          \\
                 &Baseline/320*  &15.40       &3.25                 &29.35              &79.00          \\
                 &Baseline/288*  &12.60       &2.65                 &42.39              &78.53          \\
                 &Baseline/256*  &10.00       &2.14                 &53.48              &77.21                \\
                 &HVT-S-1        &22.10       &2.40                 &47.82          &78.00           \\
                 &S$^2$ViTE      &14.60       &3.14                 &31.74          &79.22                 \\
 DeiT-S          &IA-RED$^2$     &-           &3.15                 &31.52          &79.10           \\ 
                 &DynamicViT     &22.80       &2.91                 &36.74          &79.30              \\   
                 &DynamicViT*    &22.80       &2.71                 &41.09          &79.12                \\
                 &\textbf{SPViT-320 (Ours)}   &15.44        &2.00      &56.52       &78.65             \\
                 &\textbf{SPViT (Ours)}       &22.13        &3.86      &16.09       &\textbf{79.80}             \\
                 &\textbf{SPViT (Ours)}       &22.13        &2.64      &42.61       &\textbf{79.34}             \\
 \hline 
                 &Baseline/384      &26.15       &6.55                &0.00                  &83.30      \\
      &Swin-T            &29.00       &4.50                &31.29                  &81.30    \\
                 &CvT-13/224        &20.00       &4.50                &31.29                  &81.60    \\
                 &MD-DeiT-S         &22.10       &4.60                &29.77                  &81.48    \\
                 &CPVT-Small-GAP    &23.00       &4.60                &29.77                   &81.50    \\
                 &NViT-S            &23.00       &4.70                &28.24                  &81.22    \\
 LV-ViT-S      &ATS+CvT-21        &32.00       &5.10                &22.14              &82.30    \\
                 &T2T-ViT-14        &22.00       &5.20                &20.61                  &81.50    \\
                 &CrossViT-S        &26.70       &5.60                &14.50                  &81.00    \\
                 &PVT-Medium        &44.20       &6.70                &-2.29                  &81.20    \\
                 &CoaT Mini         &10.00       &6.80                &-3.82                  &80.80    \\
                 &DynamicViT        &26.90       &4.57                &30.22              &83.00     \\
                 &\textbf{SPViT (Ours)}     &26.17       &4.28      &34.65     &\textbf{83.10}             \\
\hline 
                 &Baseline/512      &55.83       &12.67               &0.00                  &84.00          \\
     &CvT-21/224        &32.00        &7.10                &43.96               &82.50    \\
                 &DynamicViT*   &57.10        &7.35                &41.99           &83.61        \\
                 &DynamicViT    &57.10        &8.45                &33.31           &83.80         \\
                 &RegNetY-8G    &39.00        &8.00                &36.86               &81.70         \\
                 &Swin-S        &50.00        &8.70                &31.33               &83.00    \\
                 &T2T-ViT-19    &39.20        &8.90                &29.76               &81.90         \\
                 &Evo-ViT       &87.30        &9.07                &28.41               &81.11         \\
                 &VTP           &48.00        &10.00               &21.07               &80.70    \\
                 &ATS+CvT-13/384&20.00        &11.70               &7.66              &82.90         \\
LV-ViT-M         &T2T-ViT-24    &64.10        &14.10               &-11.29               &82.30         \\
                 &TNT-B         &66.00        &14.10               &-11.29                &82.80    \\
                 &Swin-B        &88.00         &15.40               &-21.55               &83.30    \\
                 &RegNetY-16G   &84.00        &16.00               &-26.28               &82.90         \\
                 &CvT-13/384    &20.00        &16.30               &-28.65               &83.00         \\
                 &ATS+CvT-21/384&32.00        &17.40               &-37.33               &83.10         \\
                 &DeiT-B        &86.60        &17.60               &-38.91               &81.80        \\
                 &CrossViT-B    &104.70       &21.20               &-67.32               &82.20    \\
                 &CvT-21/384    &32.00        &24.90               &-96.53              &83.30         \\
                 &\textbf{SPViT (Ours)}       &55.85               &7.32            &42.23      &\textbf{83.71}  \\ 
\bottomrule
\end{tabular}}
\caption{Results of different ViTs on ImageNet-1K. We compare the proposed SPViT with existing ViT pruning methods under comparable GFLOPs and the number of parameters. Note that ``*'' refers to our reproduced results to obtain models with similar GFLOPs for comparison. Negative values in FLOPs reduction mean FLOP increases. Baseline/160/192/288/384 indicates the embedding dimensions. SPViT-256/320 indicates pruning from DeiT scaling model of 256/320 embedding dimensions.}
\label{main_results}
% \vspace{-0.4cm}
\end{table*}

\end{document}